\documentclass[conference]{IEEEtran}

\usepackage[utf8]{inputenc}
\usepackage{amsfonts,amssymb,amsmath,latexsym,amsthm}
\usepackage{bbm}
\usepackage{todonotes}
\usepackage{framed}
\usepackage{caption}
\usepackage[inkscapearea=page]{svg}
\usepackage{subcaption}
\usepackage{esint} 
\usepackage{lipsum}  
\usepackage{balance}
\usepackage{courier}
\usepackage[linesnumbered,ruled]{algorithm2e}

\usepackage{xcolor}

\definecolor{forest_green}{RGB}{178,230,91}
\definecolor{aqua}{RGB}{0,222,222}
\definecolor{royal_blue}{RGB}{65,105,225}
\definecolor{brick_red}{RGB}{203,65,84}
\definecolor{burnt_orange}{RGB}{255,165,0}

\usepackage{times}
\usepackage[numbers]{natbib}
\usepackage{multicol}
\usepackage[bookmarks=true]{hyperref}

\hypersetup{
   colorlinks=true,
}

\def\Vec#1{\!\!\hbox{$#1$\kern-0.38em\lower0.85em\hbox{$\vec{}\,$}}\,}%
\newcommand{\bbm}{\begin{bmatrix}}
\newcommand{\ebm}{\end{bmatrix}}
\newcommand{\mbf}[1]{\mathbf{#1}}

\newcommand{\mbb}[1]{{\mathbb{#1}}}

\newcommand{\mc}[1]{\mathcal{#1}}

\newcommand{\eq}[1]{\begin{align}\begin{split}#1\end{split}\end{align}}

\theoremstyle{definition}
\newtheorem{definition}{Definition}
\newtheorem*{definition*}{Definition}

\begin{document}
\title{Inexact Loops in Robotics Problems}

% You will get a Paper-ID when submitting a pdf file to the conference system
% \author{Author Names Omitted for Anonymous Review. Paper-ID 12}

\author{\authorblockN{Erik Nelson}
\authorblockA{Nuro, Inc.}}

\maketitle

\begin{abstract}
Loops are pervasive in robotics problems, appearing in mapping and localization, where one is interested in finding loop closure constraints to better approximate robot poses or other estimated quantities, as well as planning and prediction, where one is interested in the homotopy classes of the space through which a robot is moving. We generalize the standard topological definition of a loop to cases where a trajectory passes close to itself, but doesn't necessarily touch, giving a definition that is more practical for real robotics problems. This relaxation leads to new and useful properties of inexact loops, such as their ability to be partitioned into topologically connected sets closely matching the concept of a ``loop closure'', and the existence of simple and nonsimple loops. Building from these ideas, we introduce several ways to measure properties and quantities of inexact loops on a trajectory, such as the trajectory's ``loop area'' and ``loop density'', and use them to compare strategies for sampling representative inexact loops to build constraints in mapping and localization problems. 
\end{abstract}

\IEEEpeerreviewmaketitle

\section{Introduction}

The study of time-parameterized trajectories is common across many subfields of robotics. Nearly all robotics applications require estimating or forecasting the state of some system through its state space, whether that state includes the joint angles of a manipulator, the location and orientation of a mobile robot, a vector of weights in a machine learning model during training, or the pose of a nearby agent. Fundamental to the study of trajectories is the concept of a loop - a trajectory whose start and end points are the same.

While the standard topological definition of a loop is extremely useful in developing algorithms and theory, it fails to suit our needs in real robotics problems for several reasons. In practice, it is rare to have certainty that a path's start and end points are identical. Sensor noise, tracking error, and sometimes deliberately added stochasticity lead to an inability to know a system's state exactly, therefore making it impossible to tell whether that state is equal to its former self. Furthermore, trajectories with loops are simply not common on many of the high dimensional state spaces considered in robotics. For example, while in 2D a discrete random walk will return to its starting point with probability one~\cite{polya1921aufgabe}, in 3D this probability is only 34\%~\cite{mccrea1940xxii}, leading to the expression ``a drunk [person] will find their way home, but a drunk bird may get lost forever''. Is is intuitive that a random continuous 1D path embedded in an $n$D manifold will self intersect less frequently with larger $n$.

Due to a definitional void, we seek to describe a relaxed type of loop that admits a non-zero distance between its start and end points. The idea of an inexact loop is not novel; it is widely understood that a loop closure~\cite{grisetti2010tutorial} in the context of Simultaneous Localization and Mapping (SLAM)~\cite{cadena2016past} refers to this idea already. However, inexact loops have not been rigorously defined in a general way that captures trajectories over arbitrary smooth manifolds, and their properties have not been thoroughly studied. An exploration of the functions built en route to a general definition uncovers many new concepts that do not apply to exact loops.

% Fig: Motivating Figure --------------------------------------------------------------
\begin{figure}[]
    \centering
    \includegraphics[width=0.4\textwidth]{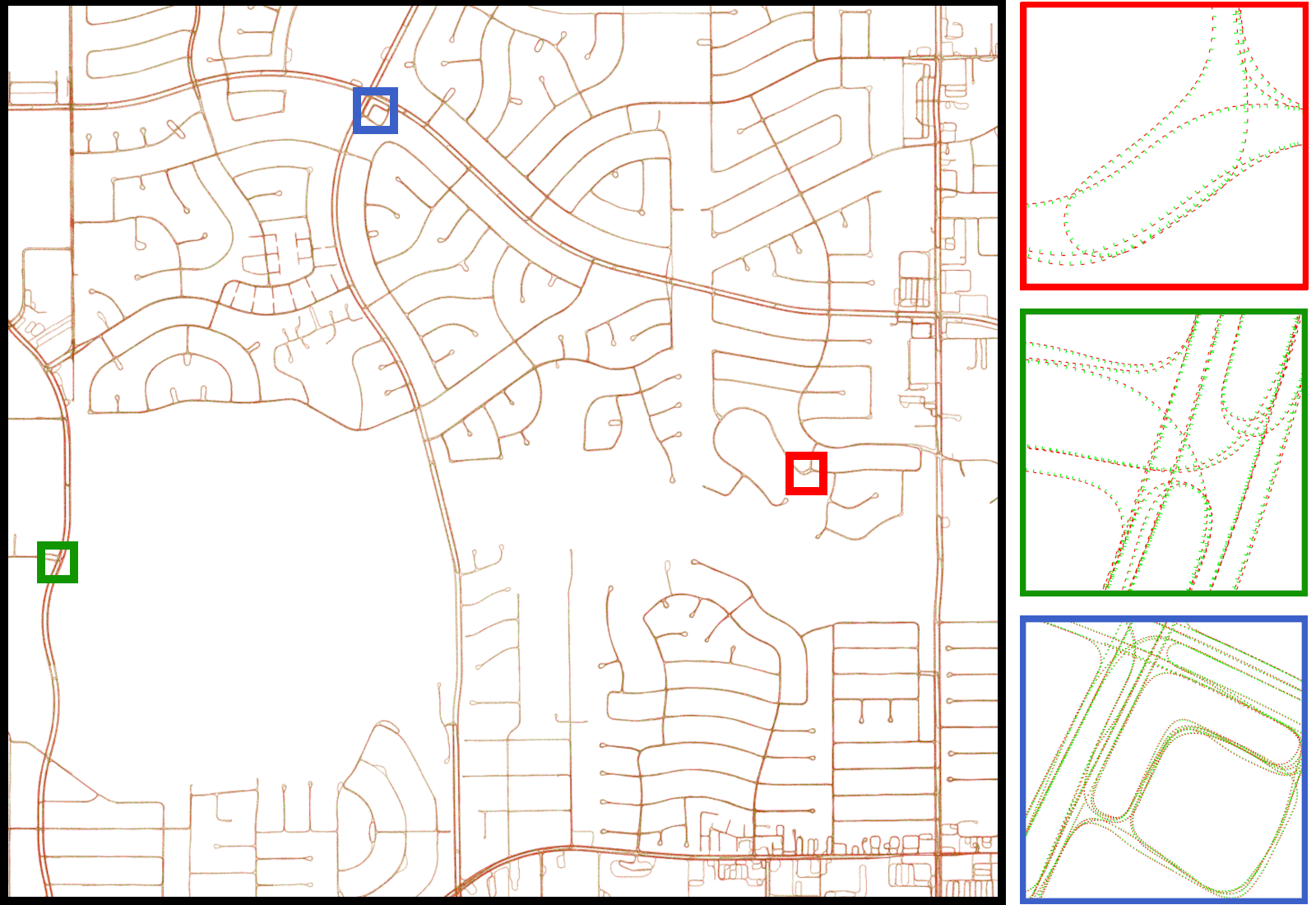}
    \caption{Characterizing loops on trajectories in the real world can be challenging. A set of trajectories spanning road surfaces in an urban area are shown. Notably, no two points on the trajectories meet exactly in $\mathbb{SE}(3)$. Which parts are ``loops''? \label{fig:motivating_example}}
\end{figure}

A primary motivator for this research is the desire to better understand how ``dense'' a trajectory is with potential loop closure constraints when solving large scale SLAM problems on billions of variables, thereby identifying locations where constraints might be over or under sampled. Improving sampling algorithms that choose which loops to build constraints out of may allow us to achieve the same quality of reconstructions in less time, which is important when the number of possible pairs of discretely sampled points on a large set of trajectories is in the trillions to quadrillions. We therefore focus our discussion and examples on the mapping and localization setting, but note that the concepts introduced here are broadly applicable to planning, prediction, and other settings as well. For example, the machinery for clustering inexact loops may be useful in optimal exploration and path coverage problems, or in sampling based planning, where one wants to know if two trajectories that do not begin or end at exactly the same location belong to the same homotopy class~\cite{bhattacharya2012topological}.

An exemplary set of reconstructed trajectories covering road surfaces in a city area are shown in Fig.~\ref{fig:motivating_example}. While it is easy to identify pairs of points that are near one another, answers to the following are less clear:
\begin{itemize}
\item Which pairs of points constitute a single ``loop''?
\item How often is a particular point near the rest of the trajectory or a particular segment?
\item How do we identify false negative detections of loops?
\item How can we build a sample of point pairs that describes cases where the trajectory leaves and returns to itself?
\item When is such a sample representative?
\end{itemize}
%Answering these questions requires a deeper understanding of inexact loops.

Our contributions are as follows. In Sec~\ref{sec:loops}, we provide a formal definition of an inexact loop and discuss its topological properties. In Sec.~\ref{sec:measuring} we introduce the concepts of loop duration, loop area, and loop density, which are measurements that can be used to analyze inexact loops on a trajectory. In Sec.~\ref{sec:sampling} we compare methods for sampling inexact loops from a trajectory, showcasing their effects on reconstructed trajectories driven by robots on urban streets.

\section{Related Work}
\label{sec:related_work}

Loops have been studied in a broad variety of topics, appearing generally in the study of algebraic topology, and specifically in any situation involving tracking the state of a system through its state space, e.g. in orbital mechanics, optimization, signal processing, quantum physics, and robotics. We split discussion of related work into the study of loops in mathematics and other fields, trajectory similarity in robotics contexts, and loops as they pertain to SLAM.

\subsection{Loops in Mathematics and Other Fields}

The formalization of loops in mathematics dates back to at least the early 19th century with the introduction of Cauchy's contour integrals~\cite{cauchy1821cours}. In 1895, Poincar\'{e} established the idea of the fundamental group of a topological space by studying classes of loops that can be continuously deformed to one another~\cite{poincare1895analysis}. This foundational idea aided in the classification of 2D closed surfaces, and ultimately nucleated the ongoing abstract study of homology~\cite{maclane2012homology}. While loops can be defined over any topological space~\cite{adams1978infinite}, inexact loops require the addition of a metric on that space, pushing them into the realm of differential geometry where they have not been studied.

Outside of pure mathematics, loops in $\mathbb{R}$ and $\mathbb{Z}$ appear in the context of periodic functions within signal processing and quantitative finance~\cite{oppenheim2001discrete, sornette2001significance}, and notions of distance on curved manifolds appear prevalently in geodesy~\cite{torge2012geodesy}, satellite tracking~\cite{chobotov2002orbital, patera2001general}, and physics~\cite{curtis1985differential}.

\subsection{Trajectory Similarity in Robotics}

There are many existing methods for comparing two trajectories, comparing a trajectory to itself, or clustering trajectories by similarity~\cite{magdy2015review, anagnostopoulos2006global, besse2015review, li2018deep}. Famous examples of these include Frech\'{e}t distance~\cite{eiter1994computing}, and the dynamic time warping algorithm~\cite{muller2007dynamic}. Computing pairwise distances between trajectory points~\cite{shao2015motion}, a technique expanded upon here, is general enough to exist as a simple function call in some software packages~\cite{pebesma2020package}. Within planning, loops and distance on manifolds arise in the study of homotopy constraint satisfaction~\cite{bhattacharya2012topological, kuffner2004effective}.

\subsection{Loop Closing in SLAM}

Loops and trajectories on curved manifolds are both heavily explored in the topic of SLAM. The identification~\cite{chen2020overlapnet,glover2012openfabmap,latif2014online} and long-term management~\cite{mazuran2016nonlinear,carlevaris2014generic,tian2018near} of loop closure constraints remain some of SLAM's hardest problems~\cite{cadena2016past}. Most introductory SLAM literature does not define the concept of a loop rigorously, and typical explanatory diagrams feature simple exact loops in $\mathbb{R}^2$, leaning on practitioners to interpret what constitutes a loop in the wild. Several works investigate the problem of clustering and sampling loop closures~\cite{olson2009recognizing,latif2013robust,zhang2016robust}, which we formalize here. Most closely aligned to this work is a recent investigation into proving the existence of loops on a given trajectory~\cite{rohou2018proving}, which makes headway into some of the concepts introduced below restricted to exact loops on trajectories in $\mathbb{R}^2$.

\section{Analysis and Properties of Loops}
\label{sec:loops}

In the following section we introduce the concept of a loop from topology, and relax it in order to define inexact loops which are more suitable for robotics problems. This definition adheres to the standard mental model of a ``loop closure'', which should in some way convey the fact that two separate parts of a trajectory were in close enough proximity to observe similar parts of the environment. However, the definition additionally unveils some interesting properties of inexact loops, such as the concept of a loop component, and the existence of simple and nonsimple inexact loops. Furthermore, this definition leads to several functions that can be used as intuitive tools for reasoning about inexact loops.

\subsection{Definitions}

% Fig: Example Loops --------------------------------------------------------------
\begin{figure*}[ht]
     \centering
     \begin{subfigure}[b]{0.24\textwidth}
         \centering
         \includegraphics[width=\textwidth]{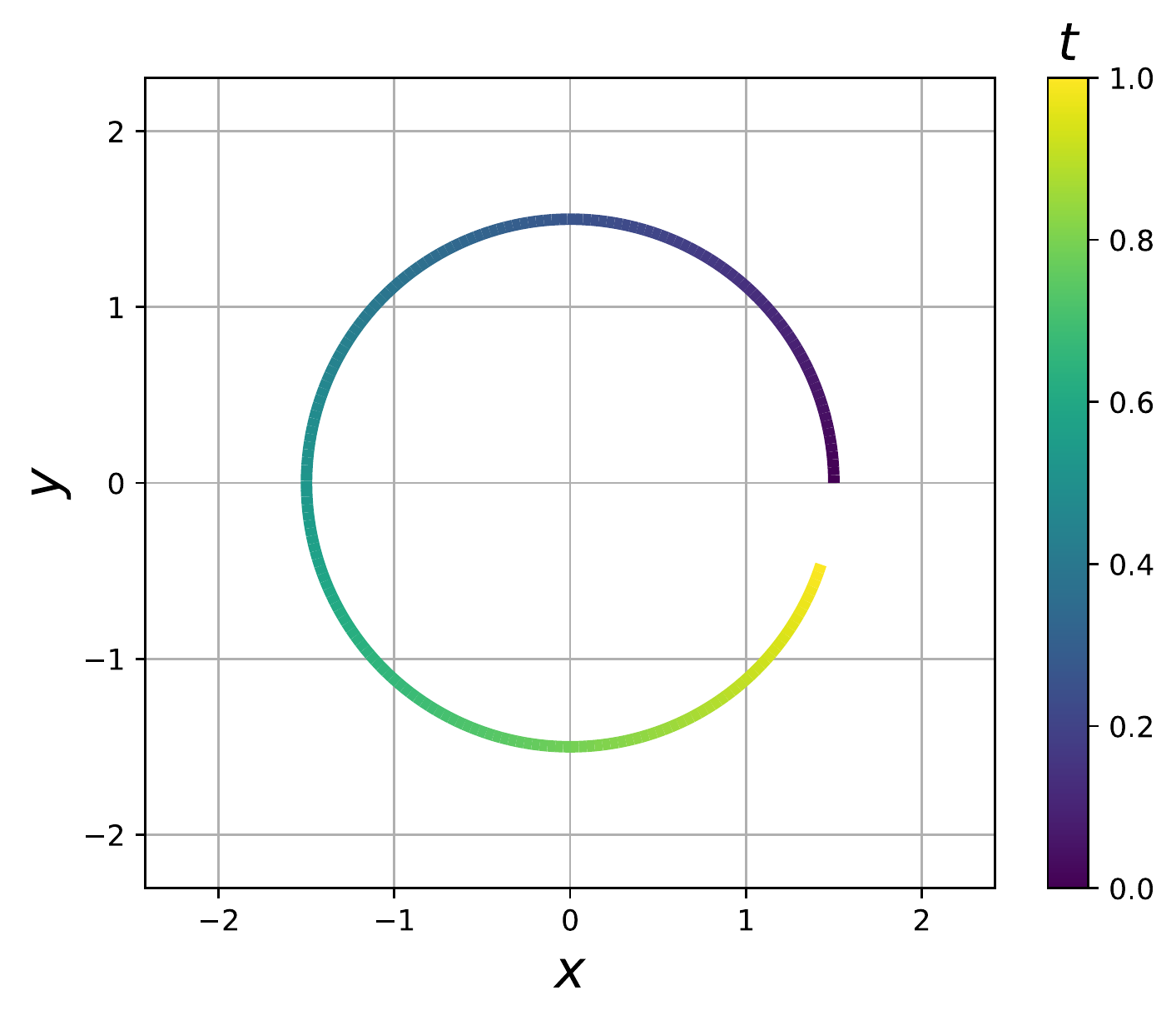}
     \end{subfigure}
     \hfill
     \begin{subfigure}[b]{0.24\textwidth}
         \centering
         \includegraphics[width=\textwidth]{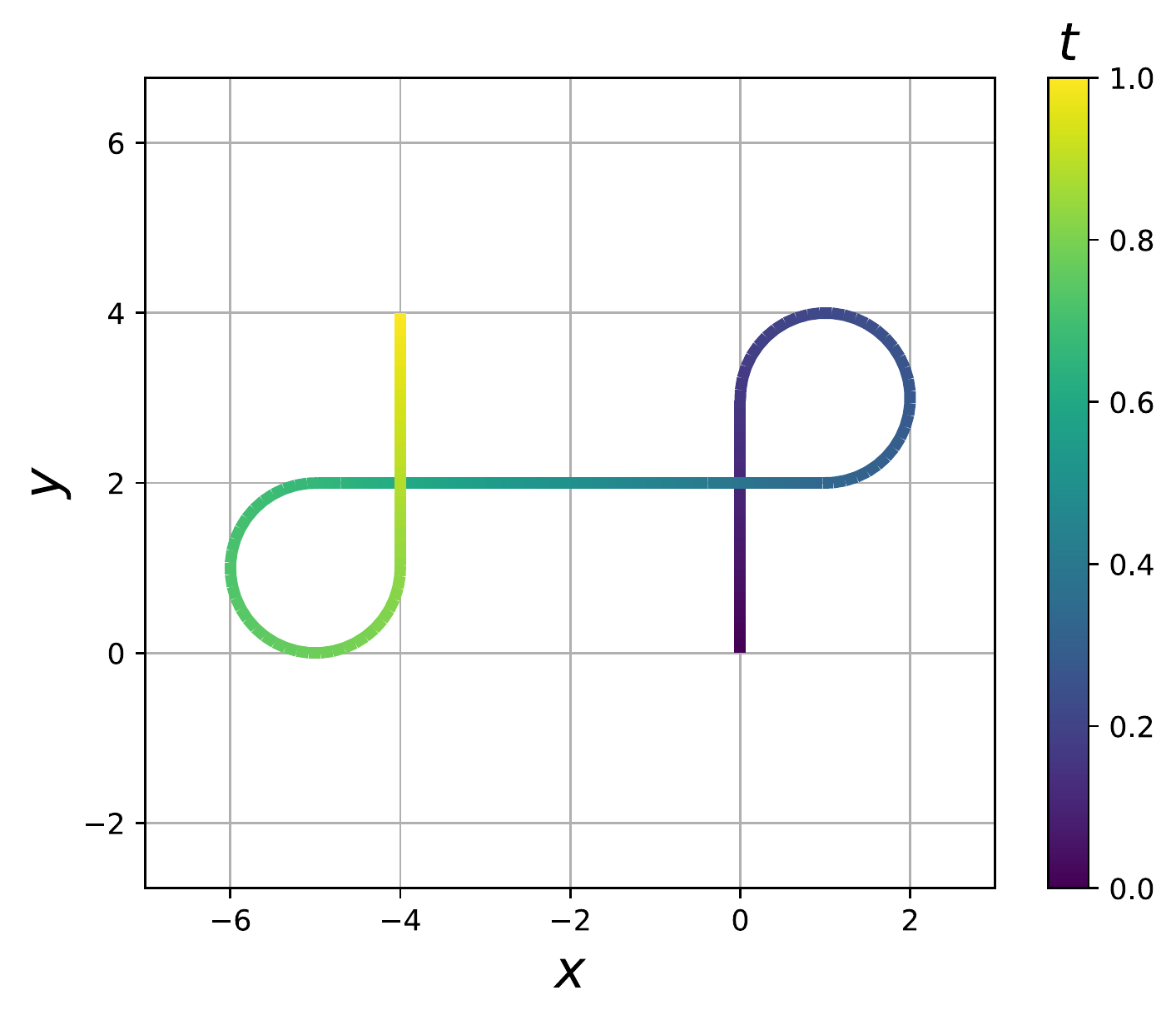}
     \end{subfigure}
     \hfill
     \begin{subfigure}[b]{0.24\textwidth}
         \centering
         \includegraphics[width=\textwidth]{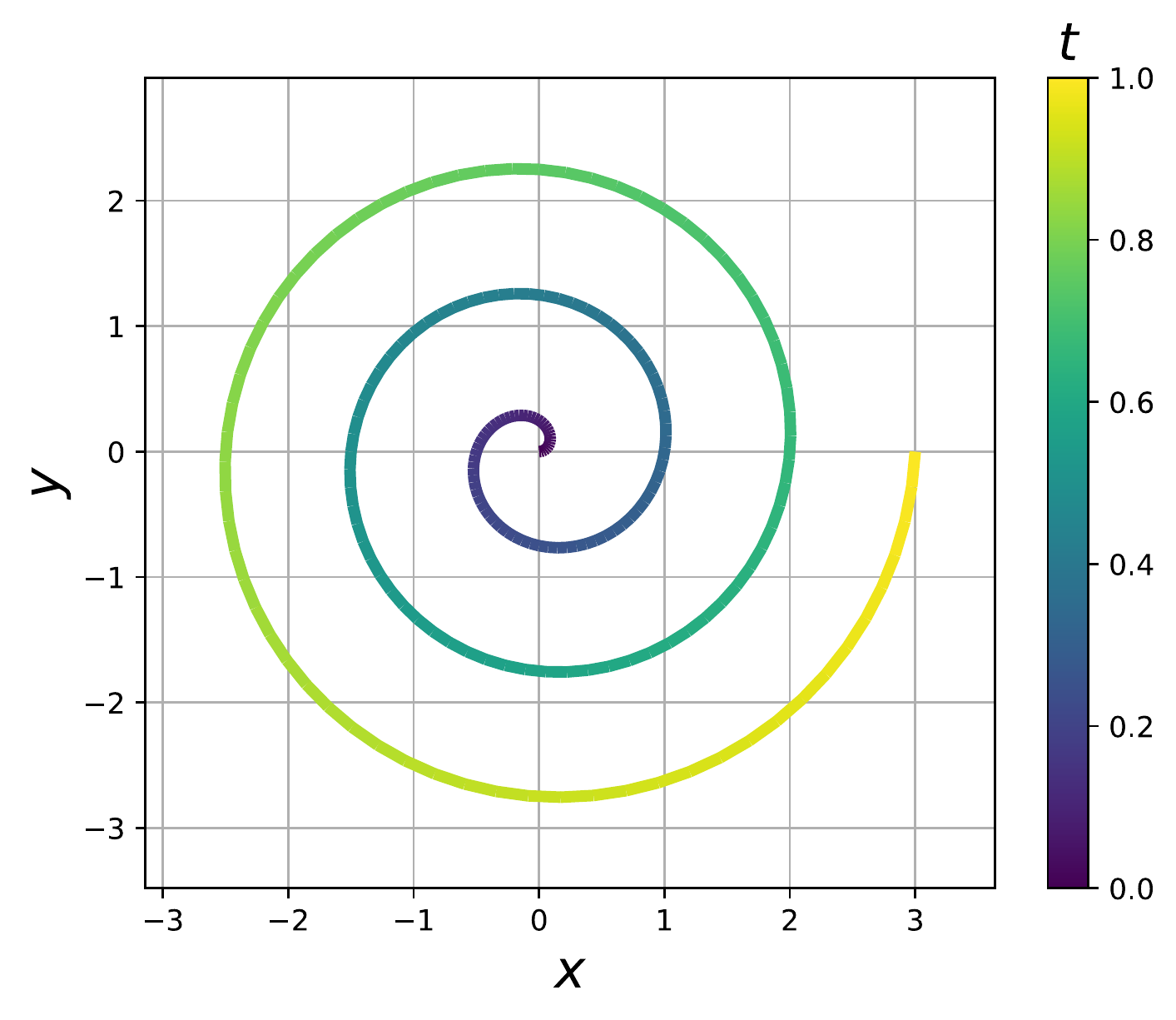}
     \end{subfigure}
     \hfill
     \begin{subfigure}[b]{0.24\textwidth}
         \centering
         \includegraphics[width=\textwidth]{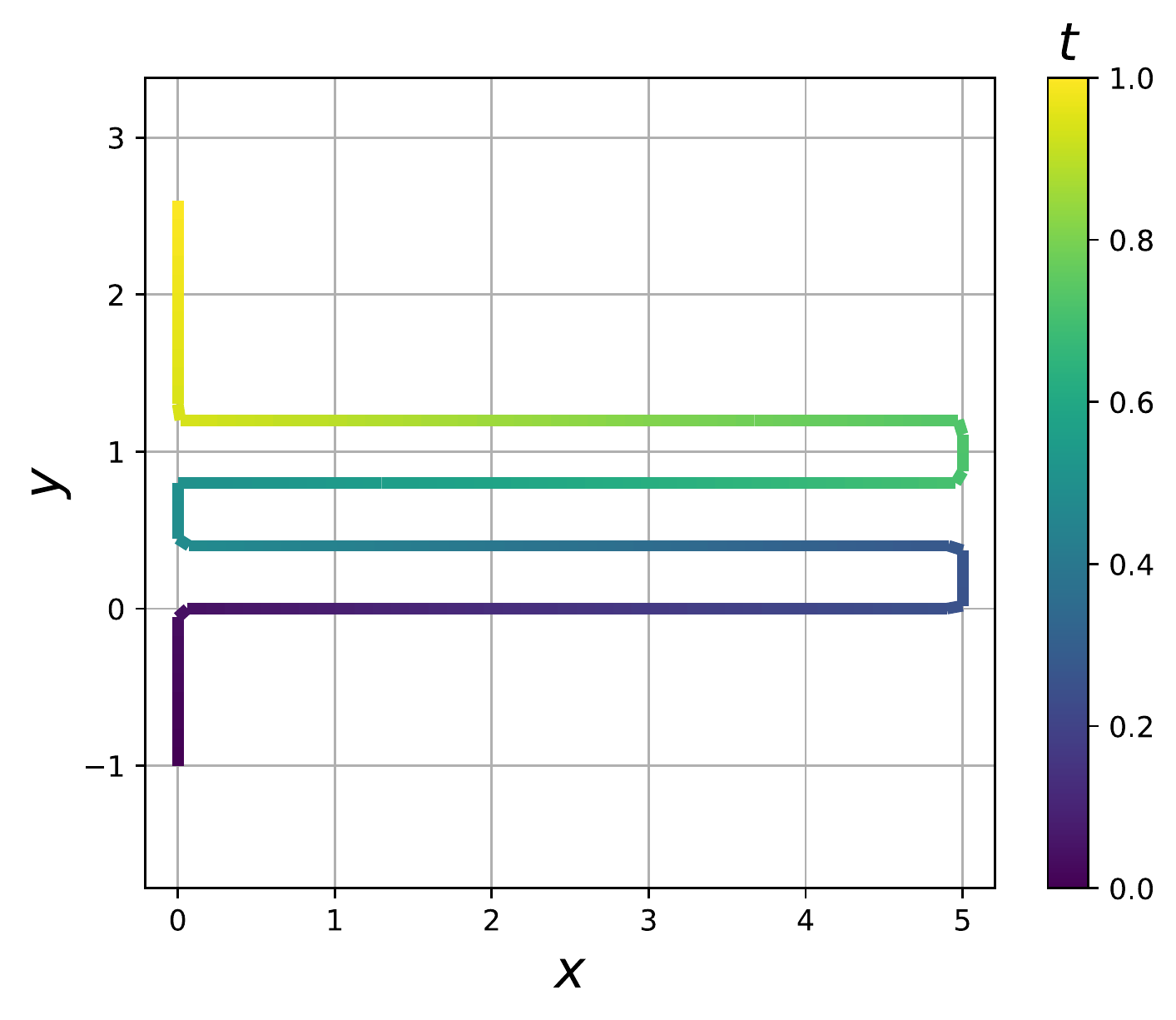}
     \end{subfigure}
     \\
    \begin{subfigure}[b]{0.24\textwidth}
         \centering
         \includegraphics[width=\textwidth]{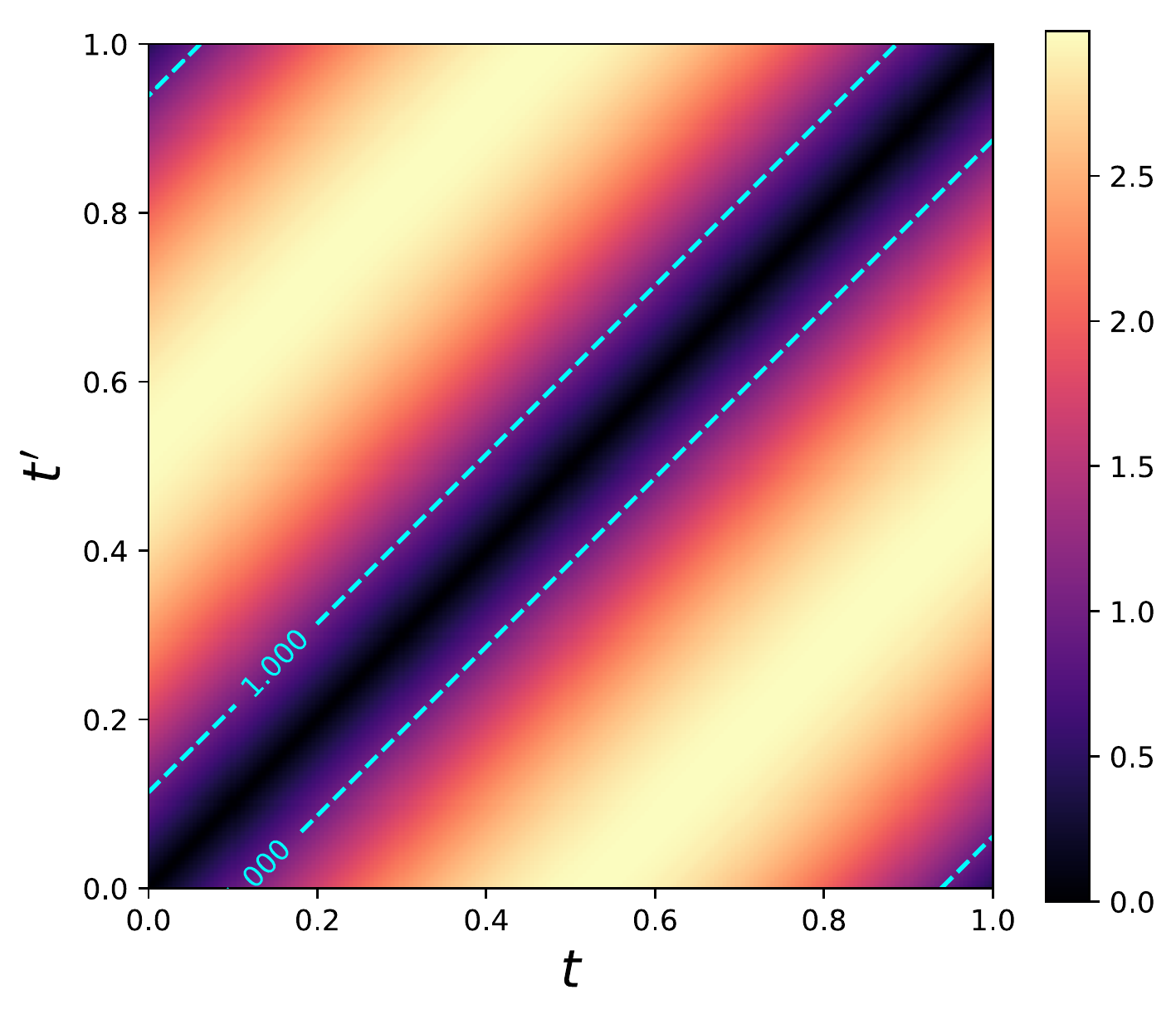}
         \caption{\label{fig:circle_example}}
     \end{subfigure}
     \hfill
     \begin{subfigure}[b]{0.24\textwidth}
         \centering
         \includegraphics[width=\textwidth]{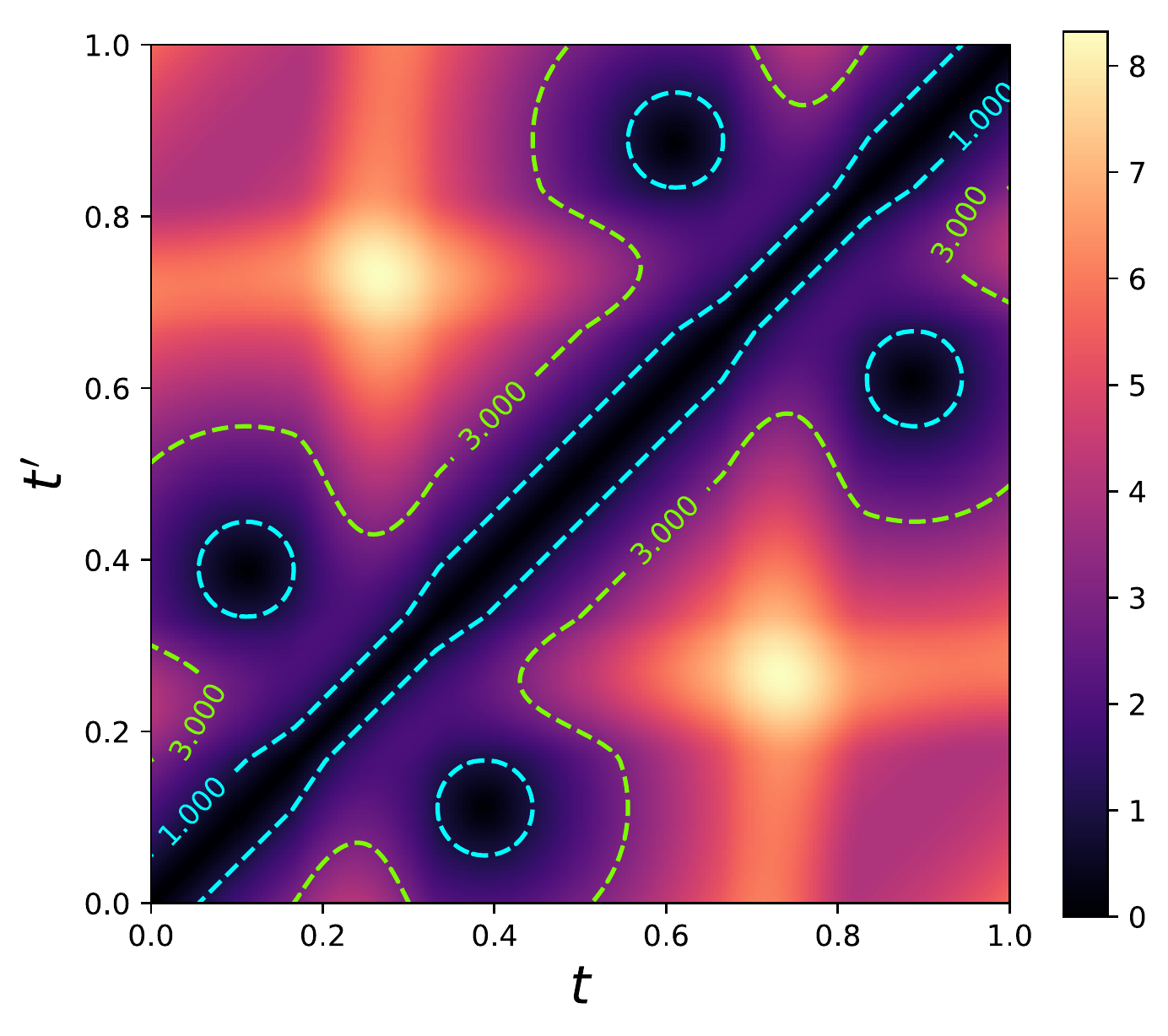}
         \caption{\label{fig:double_loop_example}}
     \end{subfigure}
     \hfill
     \begin{subfigure}[b]{0.24\textwidth}
         \centering
         \includegraphics[width=\textwidth]{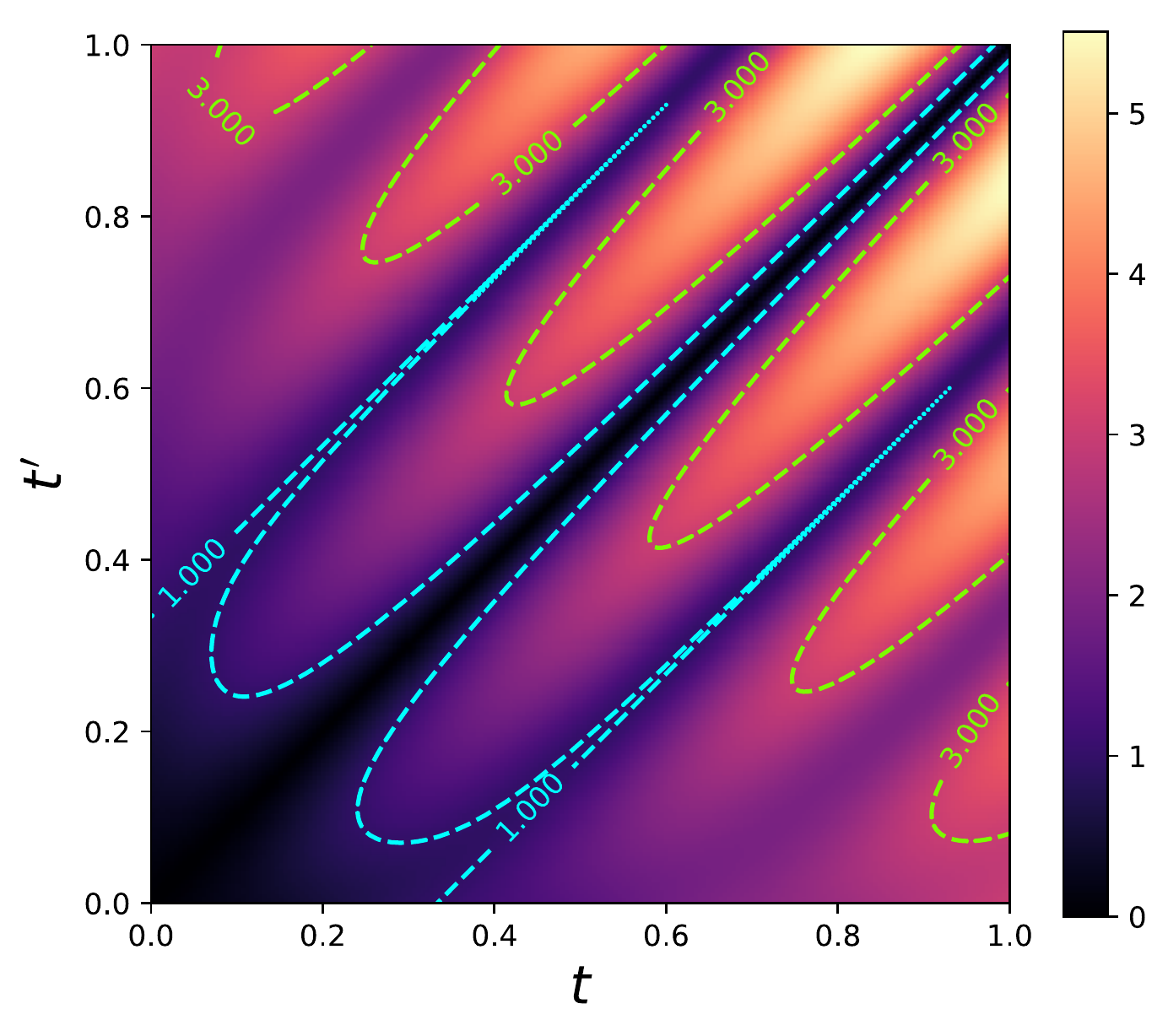}
         \caption{\label{fig:spiral_example}}
     \end{subfigure}
     \hfill
     \begin{subfigure}[b]{0.24\textwidth}
         \centering
         \includegraphics[width=\textwidth]{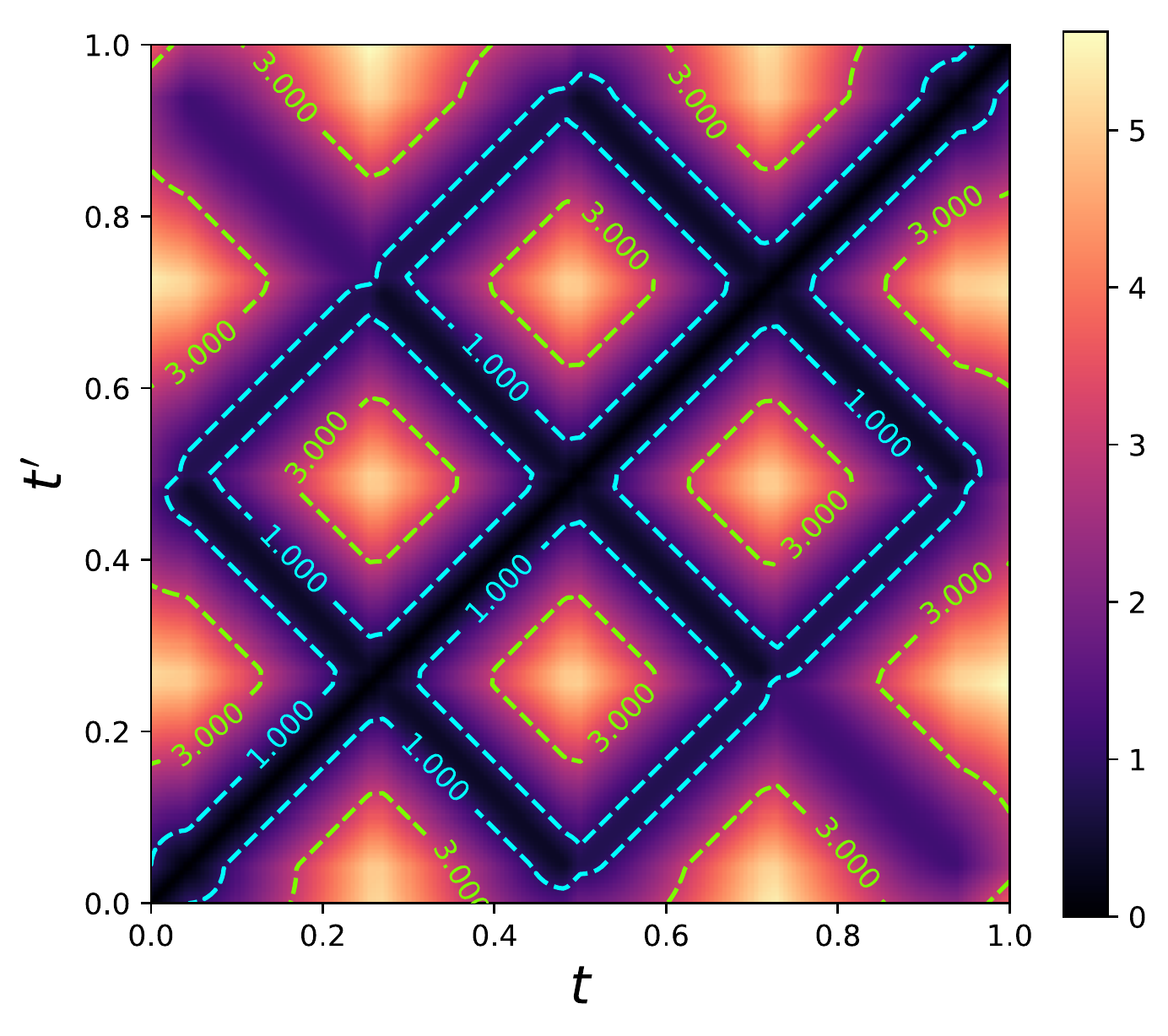}
         \caption{\label{fig:non_trivial_example}}
     \end{subfigure}
    \caption{Example trajectories $\xi : [0, 1] \rightarrow \mathbb{R}^2$ (top), and their corresponding pairwise distance functions $\pi_{\xi}(t, t')$ (bottom) under the $\mc{L}_2$ metric, with $\gamma=1$ and $\gamma=3$ isocontours highlighted. Loop components correspond to the regions below these isocontours. Fig.~\ref{fig:circle_example} displays a 95\% complete circle, highlighting the existence of the trivial loop component along the diagonal of $\pi_{\xi}(t, t')$, with one additional loop component near $(0, 1)$ (mod symmetry). Fig.~\ref{fig:double_loop_example} demonstrates that inexact loops can merge into the trivial loop component as $\gamma$ increases. Fig.~\ref{fig:spiral_example} depicts an ambiguous case, where without a rigorous definition of an inexact loop it might be difficult to identify them. Fig.~\ref{fig:non_trivial_example} displays case where the trivial loop component is multiply connected.}
    \label{fig:example_loops}
\end{figure*}

We begin by considering a continuous scalar-parameterized trajectory, $\xi : [0, 1] \rightarrow \mc{M}$, embedding a closed unit interval, which one usually interprets as an interval of time (shifted and scaled from some $[t_{0}, t_{f}]$), into a smooth manifold $\mc{M}$. Common examples of manifolds that one might encounter in robotics problems include spheres ($\mc{M}\simeq S^n$), tori ($\mc{M}\simeq S^1 \times S^1 \times \cdots$), real projective spaces ($\mc{M}\simeq\mathbb{RP}^n$), and the translations ($\mc{M}\simeq\mathbb{R}^n$), rotations ($\mc{M}\simeq\mathbb{SO}(n)$), and rigid body transformations ($\mc{M}\simeq\mathbb{SE}(n)$) of space in $n=2$ and $n=3$ dimensions. We use $\mc{E}$ to refer to the trajectory's corresponding \textit{path} - the points on $\mc{M}$ that $\xi$ passes through.
\eq{
\mc{E} = \left\{\xi(t) \,\vert\, t \in [0, 1]\right\} \subset \mc{M}.
}
Furthermore, we use $\xi([t, t']) \subseteq \mc{E}$ to refer to the subset of the path covered from $t$ to $t'$.

\begin{definition}[(Exact) Loop]
\label{def:loop}
Given a trajectory $\xi$, a \textit{loop} is a continuous function, $e : [t, t'] \subseteq [0, 1] \rightarrow \mc{E}$, such that:
\begin{itemize}
\item $\forall x \in [t, t'], \, e(x) = \xi(x)$
\item $e(t) = e(t')$
\end{itemize}
\end{definition}

We will sometimes refer to these as \textit{exact loops}. It is typical in topology for loops to be defined freely, in absence of a particular trajectory~\cite{adams1978infinite}. We instead use Definition~\ref{def:loop} to identify the loops along a provided trajectory, noting that the definitions are identical when allowed arbitrary choice of $\xi$. The topological definition of a loop enforces that the trajectory exactly meets itself at some pair of times, which is relatively rare in real robotics problems. We therefore seek to relax this definition to consider cases in which the trajectory merely comes within close proximity of itself.

%The topological definition of a loop enforces that the trajectory exactly meets itself at some pair of times, which is relatively rare in real robotics problems. For example, it is uncommon that the state space trajectory of a 7 DoF manipulator arm, or that of a camera rotating randomly in 3 DoF about a point, would ever cross themselves. We therefore seek to relax this definition to consider cases in which the trajectory merely comes within close proximity of itself.

Given a manifold $\mc{M}$, one may introduce a \textit{metric}, or distance function, $\delta : \mc{M}^2 \rightarrow \mathbb{R}_{+}$, which satisfies:
\begin{itemize}
\item $\delta(\mbf{a}, \mbf{b}) = 0 \iff \mbf{a} = \mbf{b}$ \hfill (identity)
\item $\delta(\mbf{a}, \mbf{b}) = \delta(\mbf{b}, \mbf{a})$ \hfill (symmetry)
\item $\delta(\mbf{a}, \mbf{b}) \le \delta(\mbf{a}, \mbf{c}) + \delta(\mbf{b}, \mbf{c})$ \hfill (triangle inequality)
\end{itemize}
%Intuitively, we think of a metric as a function that takes two points on the manifold and returns the distance between them. For example in $\mathbb{R}^{n}$ it is common to use the $\mc{L}_2$ norm, $\delta(\mbf{a}, \mbf{b}) = \lVert \mbf{a} - \mbf{b} \rVert_{2}$. Many possible metrics exist for $\mathbb{SO}(3)$~\cite{huynh2009metrics}, and combinations of these two can be used to define metrics on $\mathbb{SE}(3)$~\cite{zefran1996choice}.
A metric $\delta$ on $\mc{M}$ allows one to compute distances between pairs of points. Importantly, $\delta$ need not be a metric that evaluates distance between points directly, but might instead measure distance or quality of match between sensor measurements at two points~\cite{di1999distance,wang2017deep}. For a particular trajectory, we define the pairwise distance function $\pi_{\xi} : [0, 1]^2 \rightarrow \mathbb{R}_{+}$, which evaluates the metric at a given pair of times:
\eq{
\pi_{\xi}(t, t') = \delta(\xi(t), \xi(t')).
}
% We will see shortly that $\pi_{\xi}$ is a key tool in reasoning about inexact loops. 
$\pi_{\xi}$ inherits the symmetry property of $\delta$, hence $\pi_{\xi}(t, t') = \pi_{\xi}(t', t)$. Examples of trajectories on $\mbb{R}^2$ and their pairwise distance functions under the $\mc{L}_{2}$ metric are shown in Fig.~\ref{fig:example_loops}. We define the $\gamma$-sublevel set of a real-valued function $f$ as
\eq{
 \mc{S}_{\gamma}(f) = \left\{ \mbf{x} \,\vert\, f(\mbf{x}) \le \gamma, \,\, \gamma \in \mathbb{R} \right\},
}
and consider the set $\mc{S}_{\gamma}(\pi_{\xi}) \subseteq [0, 1]^2$, which contains all pairs of times for which the trajectory came within $\gamma$ distance of itself. For brevity we will refer to this set as $\mc{S}_{\gamma}$, remembering that it references a particular metric and trajectory. Using these pieces, we are ready to define inexact loops.

\begin{definition}[Inexact Loop]
\label{def:inexact_loop}
Given a trajectory $\xi$, metric $\delta$, and distance threshold $\gamma \in \mathbb{R}_{+}$, an \textit{inexact loop} is a continuous function, $l : [t, t'] \subseteq [0, 1] \rightarrow \mc{E}$, such that:
\begin{itemize}
\item $\forall x \in [t, t'], \, l(x) = \xi(x)$.
\item $(t, t') \in \mc{S}_{\gamma}$.
\end{itemize}
\end{definition}

Inexact loops can be thought of as segments of the trajectory whose start and end points are within $\gamma$ distance from one another. Definition~\ref{def:inexact_loop} is nearly identical to Definition~\ref{def:loop}, except that instead of requiring $\pi_{\xi}(t, t') = 0$, we require $\pi_{\xi}(t, t') \le \gamma$. This grants the additional flexibility to consider cases where a trajectory came close to itself, yet dissolves to the standard loop definition when $\gamma = 0$. For a given trajectory, inexact loops (which are functions), can be bijectively mapped to $\mc{S}_{\gamma}$ via $l \mapsto (\inf(\text{dom}(l)), \, \sup(\text{dom}(l)) = (t, t')$, so for notational convenience, we refer to $l$ as the 2D point $(t, t')$.

Next, we consider the partition of $\mc{S}_{\gamma}$ into a disjoint union of $k$ connected components:
\eq{ 
  \mc{S}_{\gamma} &= \bigsqcup_{i=1}^{k} \Omega_i,
}
where two inexact loops, $l$ and $l'$, belong to the same component $\Omega$ if their time pairs are path-connected in $\mc{S}_{\gamma}$. In this case we say $l \sim l'$, and can describe the component as the equivalence class $\Omega = [l] = [l']$.

Each component is a compact set of pairs of times in which the trajectory was within $\gamma$ distance of itself. For exact loops (i.e. when $\gamma = 0$), the component $[e]$ is almost always a singleton composed of one point - the loop $e$ itself, and is therefore not interesting ($[e]$ and $e$ are not conceptually distinct). In rare cases the components can be 1D or 2D compact subsets of $[0, 1]^2$, e.g. when the trajectory stops at exactly the point of intersection before continuing, or in the special case of $[(t, t)]$. By contrast, for inexact loops, the components are only singletons when the trajectory glances itself at a distance of exactly $\gamma$ without stopping. Unlike exact loops, inexact loops are conceptually distinct from the components that they belong to, and so we give these components a name.

\begin{definition}[Loop Component]
\label{def:loop_component}
Given a trajectory $\xi$, metric $\delta$, and distance threshold $\gamma \in \mathbb{R}_{+}$, a \textit{loop component} $\Omega \subseteq [0, 1]^2$ is a path-connected component of $\mc{S}_{\gamma}$.
\end{definition}

In other words, a loop component is a compact path-connected set of inexact loops. Its path-connectedness can be pictured by imagining two points on a trajectory sliding independently. Two inexact loops belong to the same loop component if there exists a way of continuously sliding the endpoints from one configuration to the other without the pair of points separating by a distance more than $\gamma$ (under the metric $\delta$).

Loop components are useful because they give us a way to enumerate the distinct locations where a trajectory has inexact loops. The concept of a loop component is more in line with intuition and loose terminology used in mapping and localization settings. We are typically more interested in finding the set of loop components for a given trajectory, than in finding the set of all loops, or samples thereof. The former contains more information than the latter, since the set of loop components is a partition of the set of all inexact loops.

\subsection{Properties of Inexact Loops and Loop Components}

We now discuss some properties of inexact loops.

Due to the symmetry of $\pi_{\xi}$ and $\delta$, we should expect to find inexact loops in pairs. The simple mental picture of sliding two independent points along a trajectory makes this symmetry clear: fixing endpoint $\xi(t)$ at a location and sliding endpoint $\xi(t')$ leads to the same distance values as fixing endpoint $\xi(t')$ at the same location and sliding endpoint $\xi(t)$. In other words, for a given inexact loop, one can form a different loop by flipping its direction. This can be observed in the plots of $\pi_{\xi}$ in Fig.~\ref{fig:example_loops}. %When this symmetry is unneeded, we can ignore it by considering only $\mc{S}_{\gamma} / \sim_{s}$, where for two loops $l = (a, b)$ and $l' = (c, d)$, $l \sim_{s} l'$ when $a = d$ and $b = c$. 

The identity property of $\delta$ implies that for any choice of $\gamma$ on any trajectory, the pair $(t, t)$ is always an exact loop, and therefore belongs to some loop component. We refer to the loop component $[(t, t)]$ as the \textit{trivial loop component}, and refer to loops within it as \textit{simple loops} (so as not to be confused with trivial loops in topology). The trivial loop component contains both exact and inexact loops whose endpoints can be continuously slid along the trajectory until they match (in both $\mc{M}$ and $[0, 1]$), without moving further than $\gamma$ from one another. Fig.~\ref{fig:spiral_example} displays a case in which for both $\gamma=1$ and $\gamma=3$, all inexact loops are simple. Indeed, one can imagine sliding the points $\xi(t)$ and $\xi(t')$ from any starting configuration in which $\pi_{\xi}(t, t') \le 1$ to meet at a single point without ever moving $1$ unit away from one another.

% Fig: Loops on Curved Manifolds --------------------------------------------------------------
\begin{figure}[t]
     \centering
     \begin{subfigure}[b]{0.24\textwidth}
         \centering
         \includegraphics[width=\textwidth]{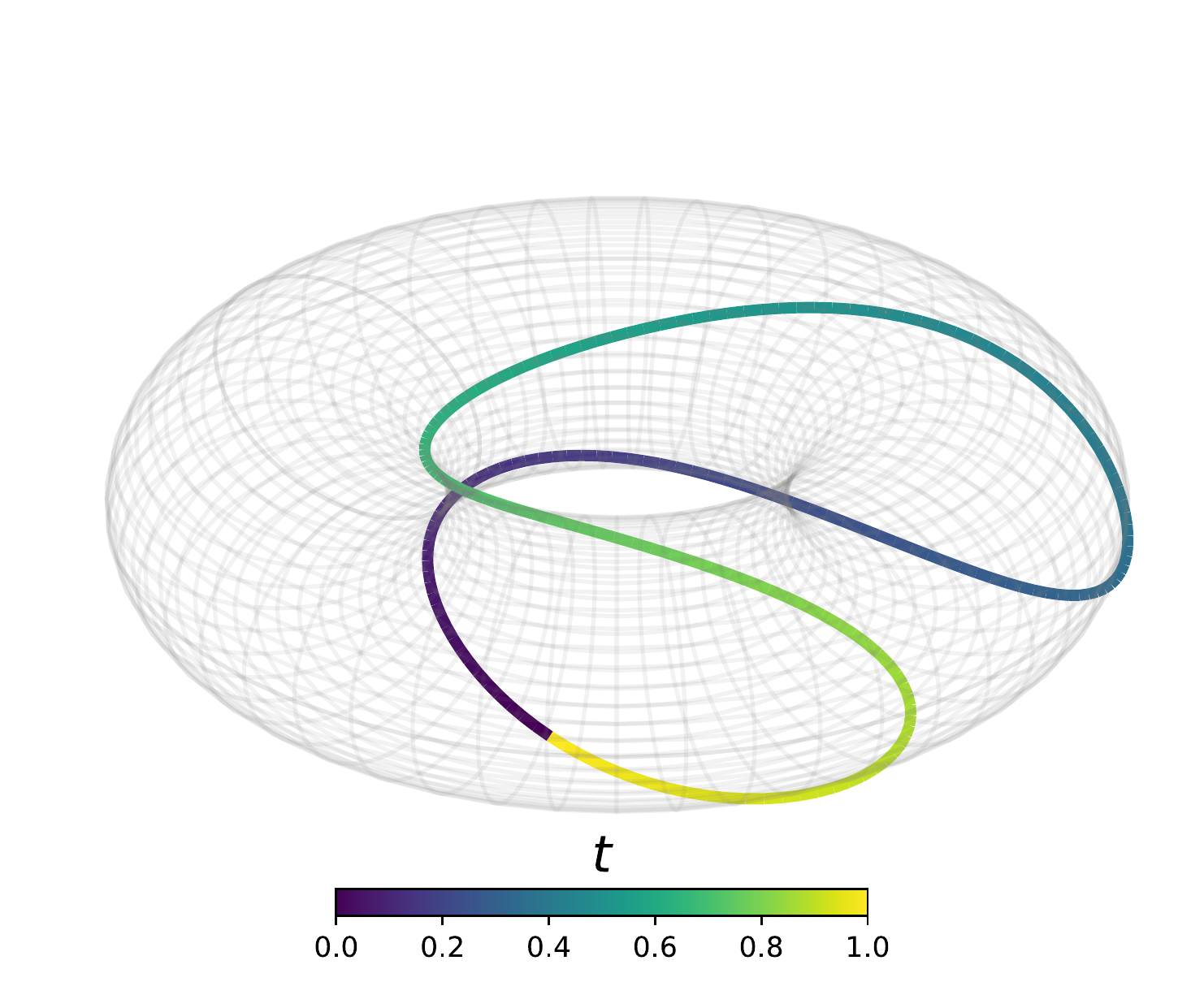}
         \caption{A circle in $S^1 \times S^1$.\label{fig:torus_3d}}
     \end{subfigure}
     \hfill
     \begin{subfigure}[b]{0.24\textwidth}
         \centering
         \includegraphics[width=\textwidth]{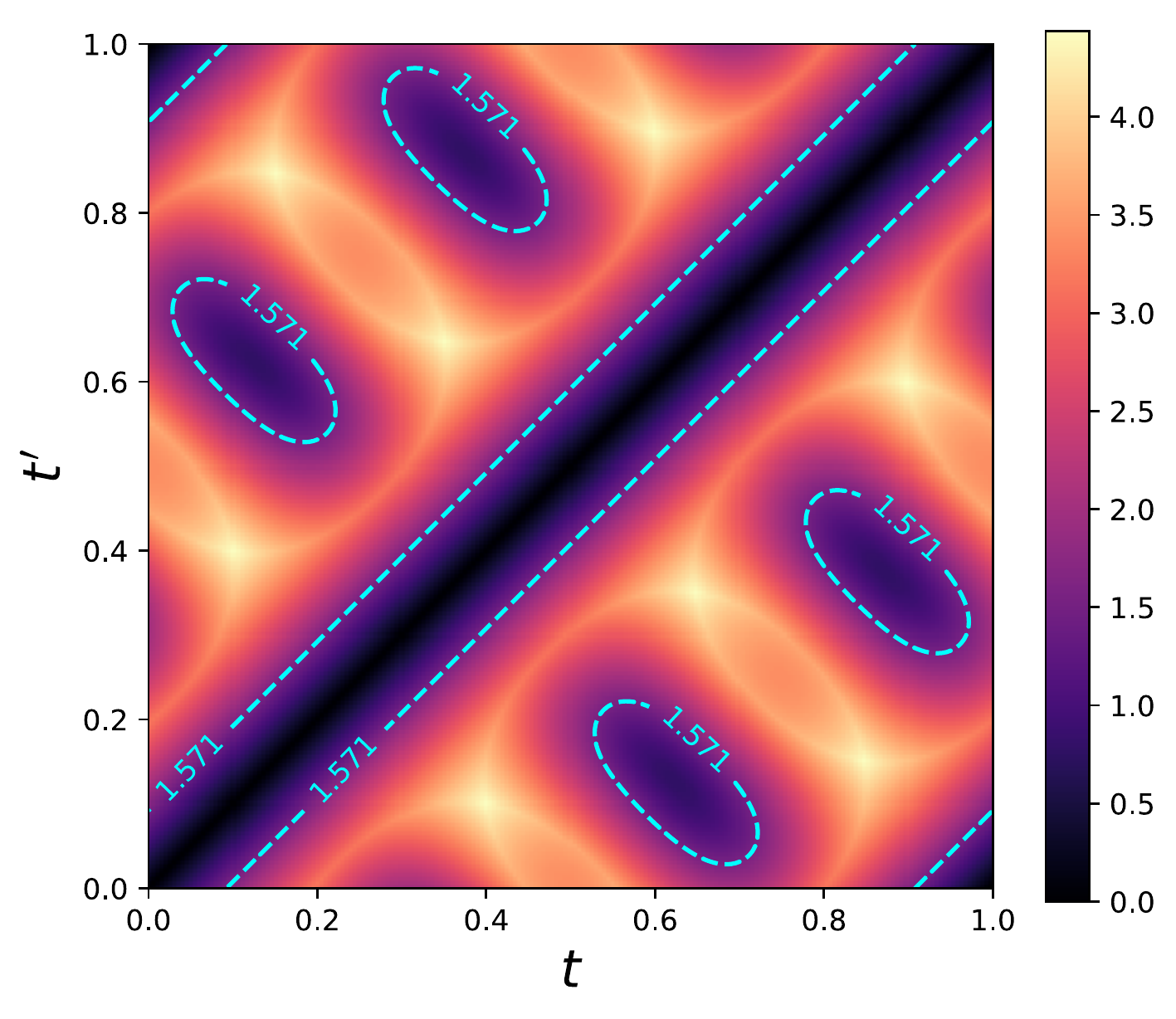}
         \caption{$\pi_{\xi}(t, t')$\label{fig:torus_dist}}
     \end{subfigure}
     \\
     \begin{subfigure}[b]{0.24\textwidth}
         \centering
         \includegraphics[width=\textwidth]{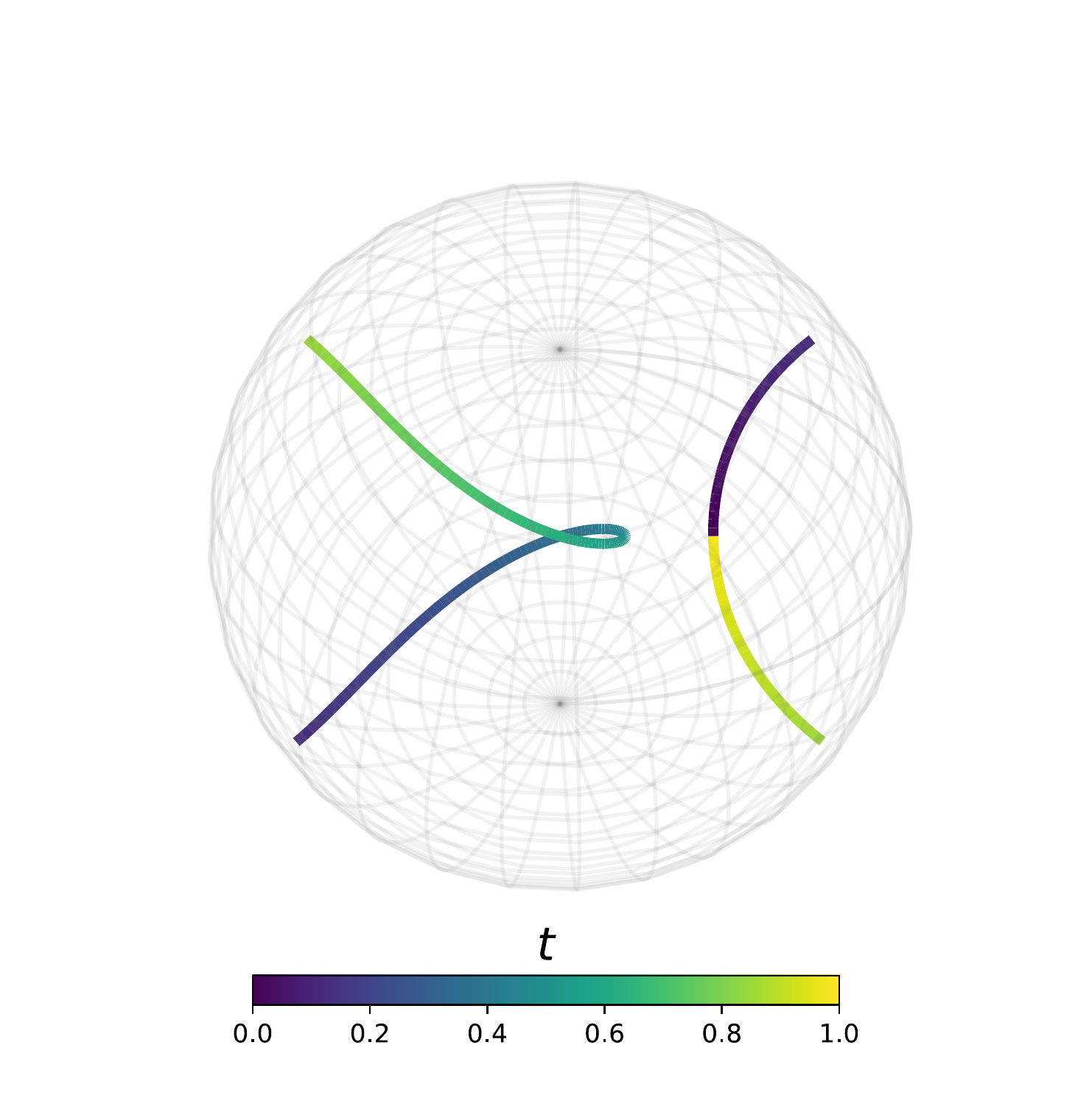}
         \caption{A circle in $\mbb{SO}(3)$.\label{fig:so3}}
     \end{subfigure}
     \hfill
     \begin{subfigure}[b]{0.23\textwidth}
         \centering
         \includegraphics[width=\textwidth]{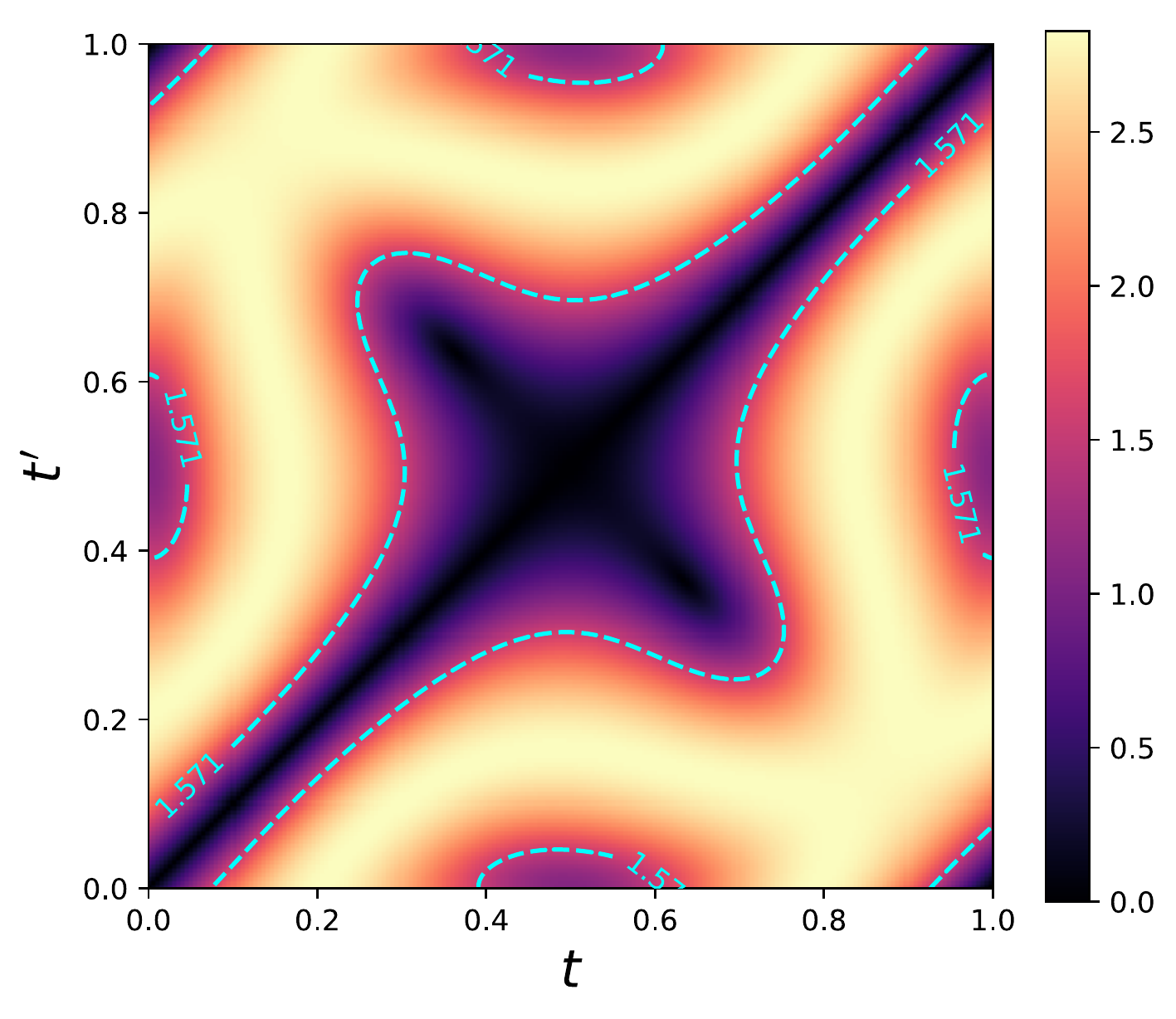}
         \caption{$\pi_{\xi}(t, t')$\label{fig:so3_dist}}
     \end{subfigure}
     \caption{Examples of circular trajectories with radius $\frac{7}{8}\pi$ on two curved manifolds, which exhibit different loop components (a third circular trajectory in $\mathbb{R}^2$ is shown in Fig.~\ref{fig:circle_example} for comparison). In $S^1 \times S^1$, we use the metric $\delta((\theta, \phi), (\theta', \phi')) = \lVert (\Delta(\theta, \theta'), \Delta(\phi, \phi')) \rVert_{2}$, where $\Delta$ returns least angular distance. In $\mbb{SO}(3)$ we use the metric $\delta(\mbf{p}, \mbf{q}) = \lVert I - \exp(\mbf{p}^{\wedge}) \exp(\mbf{q}^{\wedge})^{\top}\rVert_{F}$, where points on the path are axis-angle vectors of magnitude $\le \pi$.\label{fig:manifold_examples}}
\end{figure}

The sizes, shapes, and number of loop components are controlled by several variables. The parameter $\gamma$ clearly plays a role in deciding these quantities. For any metric, as $\gamma \rightarrow \infty$, the trivial component $[(t, t)]$ approaches $[0, 1]^2$, effectively merging with any other components. Fig.~\ref{fig:double_loop_example} displays an example in which increasing $\gamma$ from $1 \rightarrow 3$ causes all loop components to merge with the trivial component. These quantities additionally depend both on the manifold that the path is embedded in, as well as the path's location when the manifold is curved. Fig.~\ref{fig:manifold_examples} displays two trajectories whose paths carve out circles in the parameter space of two curved manifolds. On the torus $S^1 \times S^1$, the circle exactly meets itself at its start and end points, and in the process comes within close vicinity of itself two other times. On $\mathbb{SO}(3)$, the circle approaches its own start and end points at their antipodal point.

\begin{figure}[t]
     \centering
     \begin{subfigure}[b]{0.34\textwidth}
         \centering
         \includegraphics[width=\textwidth]{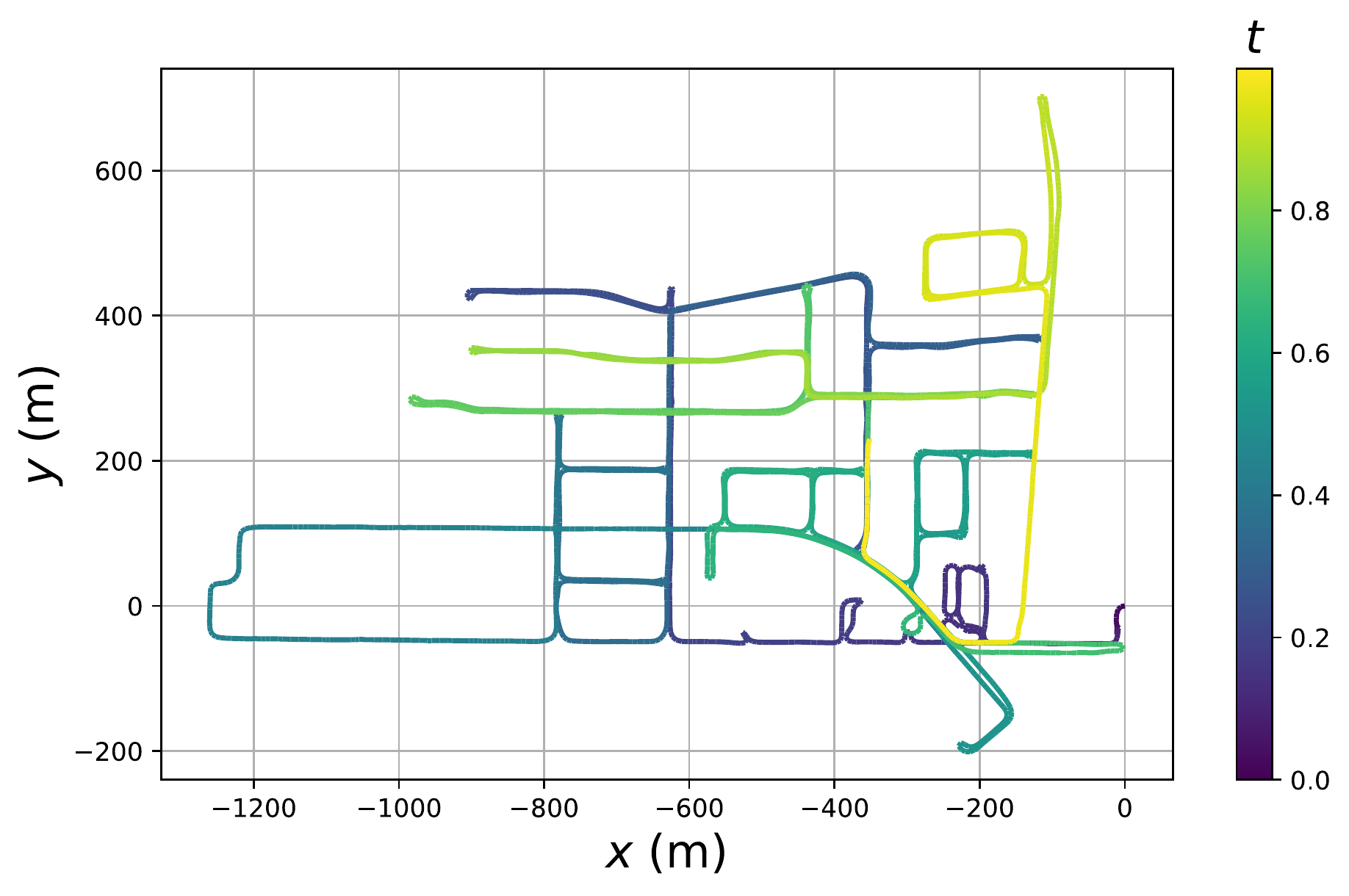}
     \end{subfigure}
     \\
     \begin{subfigure}[b]{0.24\textwidth}
         \centering
         \includegraphics[width=\textwidth]{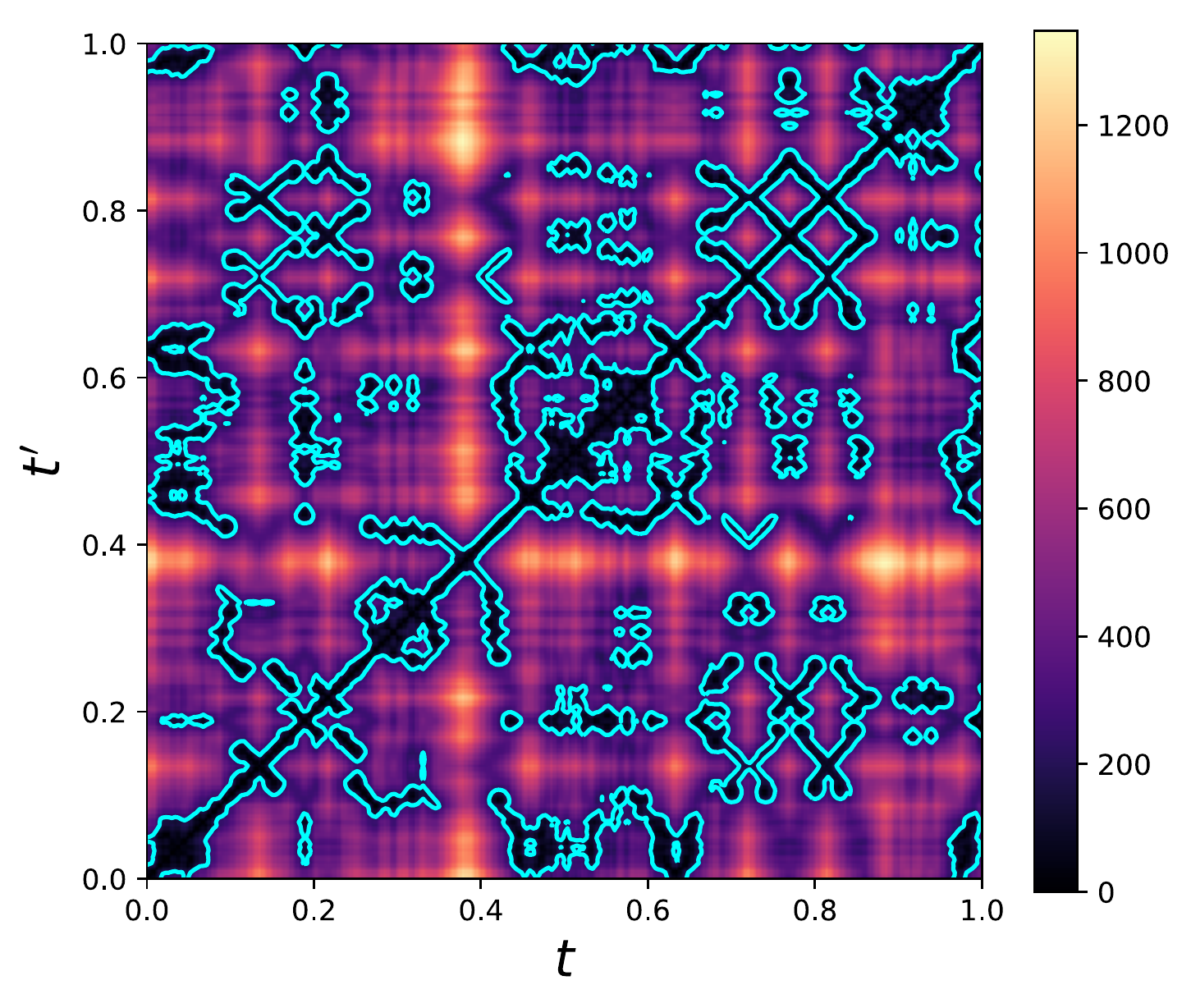}
         \caption{$\pi_{\xi}(t, t')$, $\mc{L}_{2}$ metric.\label{fig:city_traj_l2_dist}}
     \end{subfigure}
     \hfill
     \begin{subfigure}[b]{0.24\textwidth}
         \centering
         \includegraphics[width=\textwidth]{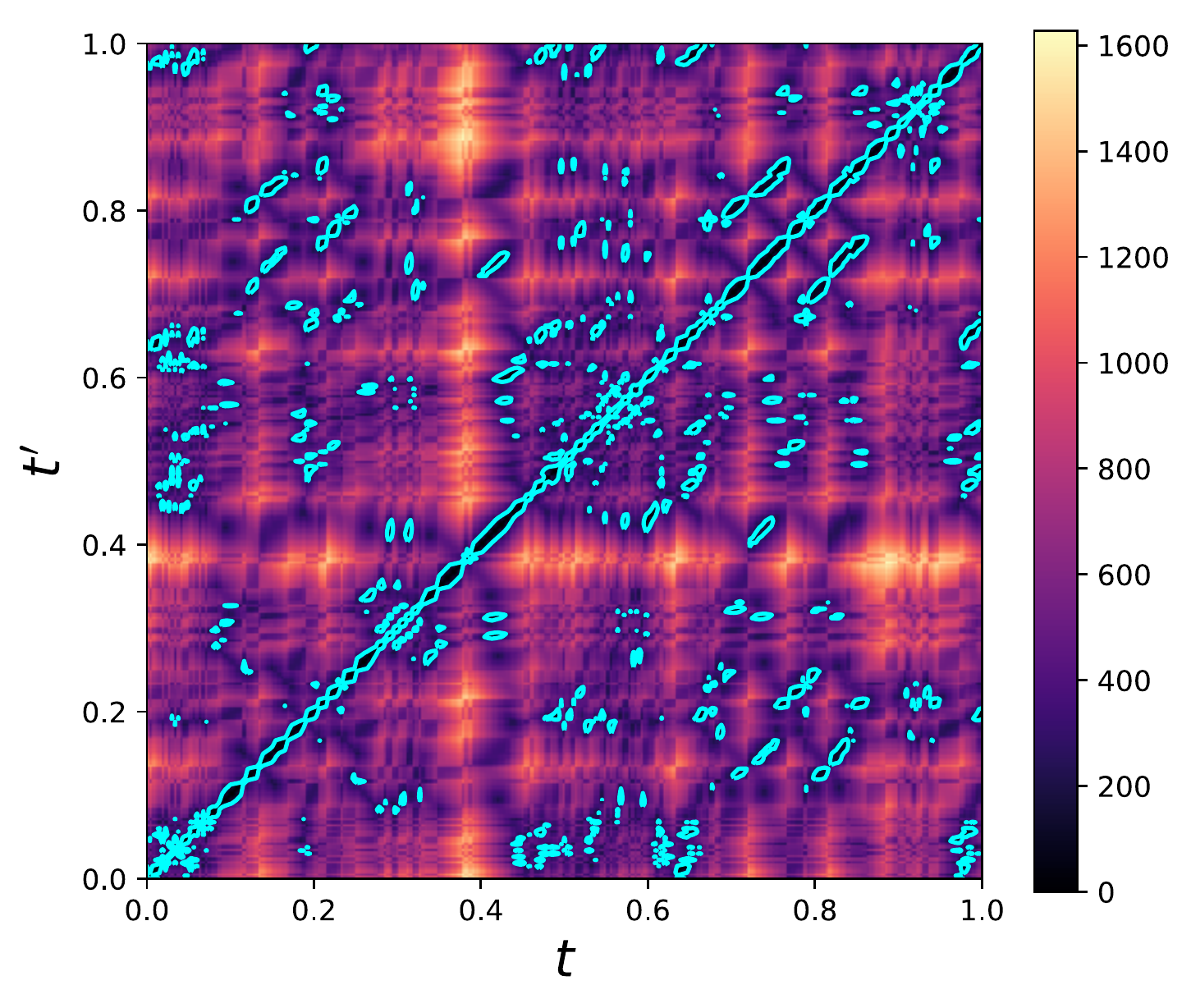}
         \caption{$\pi_{\xi}(t, t')$, $\mathbb{SE}(3)$ metric.\label{fig:city_traj_se3_dist}}
     \end{subfigure}
     \caption{A trajectory containing numerous inexact loops, with $\gamma = 200$ m isocontours highlighted for two metrics.\label{fig:real_trajectory}}
\end{figure}

Surprisingly, loop components need not be simply connected. This property is exhibited by the trivial loop component for the trajectory depicted in Fig.~\ref{fig:non_trivial_example} (note that some exact loops on the space $\mc{S}_{\gamma}$ itself cannot be shrunk to a point). This observation brings up the confusing discussion of the possible exact loops within the set of a trajectory's inexact loops. Because loop components are closed path-connected topological subspaces of $[0, 1]^2$, they are homeomorphic to the 2D disk punctured at $n$ points. This implies that the fundamental group of a loop component itself can only be the free group on $n$ generators for some $n$, which is rather uninteresting. Anecdotally, data from real trajectories implies that it is common to encounter cases with $n > 0$. Fig.~\ref{fig:real_trajectory} shows a real trajectory taken on $\mbb{SE}(3)$, where some loop components are multiply connected under the $\mc{L}_2$ metric on translations, but not under an $\mbb{SE}(3)$ metric that incorporates rotation.

% Fig: Real trajectory distances --------------------------------------------------------------

Given a loop component $\Omega$, we can consider the set of 2D points along its boundary $\partial \Omega$. This boundary traces out paths taken by pairs of points on the trajectory that stay a fixed distance of $\gamma$ from one another. To visualize this, imagine a rigid bar of length $\gamma$ connecting two points on the trajectory as they slide together. Similarly, for any closed subset $\Lambda \subset [0, 1]^2$, we can consider the boundary $\partial \Lambda$. Previous work has shown that one can identify exact loops by finding points in $\mc{S}_{0}$ about which such a boundary has a non-zero winding number~\cite{rohou2018proving}. The same can be said for inexact loops in $\mc{S}_{\gamma}$.

\section{Measuring Inexact Loops on Trajectories}
\label{sec:measuring}

%\begin{table*}[t]
%\begin{center}
%\caption{Inexact Loop Concepts.\label{table:concepts}}
%\begin{tabular}{|l|l|l|} 
% \hline
% Concept & Symbols & Description \\
% \hline
% Inexact Loop & $l$, $(t, t')$ & A trajectory segment whose endpoint $\xi(t')$ is within a distance $\gamma$ of its starting point $\xi(t)$. \\ 
% Loop Component & $\Omega, [(t, t')]$ & A compact, path-connected set of inexact loops; a subset of all inexact loops on a trajectory. \\ 
% Trivial Loop Component & $[(t, t)]$ & The unique loop component that includes all (exact) loops of the form $(t, t)$. \\
% Simple Loop & none & A loop belonging to the trivial loop component. Not necessarily of the form $(t, t)$. \\
% Loop Boundary & $\partial \Omega$ & The 1D boundary of a loop component, describing times for which $\xi(t)$ is exactly $\gamma$ from $\xi(t')$. \\
% Loop Duration & $\tau_{\mc{X}}(t)$ & The duration for which a subset of a trajectory is within $\gamma$ distance of the point $\xi(t)$. \\ 
% Loop Area & $\alpha_{\mc{X}}(t, t')$ & A measure of how many inexact loops begin or end within the segment $\xi([t, t'])$.  \\
% Loop Density & $\rho_{\mc{X}}(t, t')$ & The average duration for which a point on $\xi([t, t'])$ is within a distance $\gamma$ of other points. \\
% \hline
%\end{tabular}
%\end{center}
%\end{table*}

It would be useful to have measurements that describe various loop-related properties of a trajectory, such as how many inexact loops begin or terminate within a particular segment, or how many inexact loops are connected to the average point. For example, in SLAM problems in which an approximate absolute initial estimate is available (e.g. from GNSS, or gravity readings), these measurements would help identify cases when loop closure constraints that should have been detected are missed, simply by examining the trajectory. False negatives of this nature can lead to large errors and should be avoided at all costs. Alternatively, these measurements could be used to determine when a particular loop component has too many loop closure constraints, leading to overconfidence in an estimated reconstruction of the trajectory.

To build towards these measurements, we consider measuring the total duration for which other parts of the trajectory are within a distance $\gamma$ of a point fixed at time $t$.
\eq{
  \tau_{\mc{X}}(t) &= \int_{0}^{1} \mbf{1}_{\mc{X}}(t, t') \, dt'.
  \label{eq:loop_duration}
}
Here, $\mbf{1}_{\mc{X}} : X \rightarrow \{0, 1\}$ is the indicator function, returning $1$ when its argument is inside the set $\mc{X}$, and $0$ otherwise. Depending on the subset $\mc{X} \subseteq [0, 1]^2$ in consideration, $\tau_{\mc{X}}$ can be used to measure the duration of all nearby points on the trajectory ($\mc{X}=\mc{S}_{\gamma}$), or just those falling within particular loop components ($\mc{X}= \Omega_1 \cup \Omega_2 \cup \cdots$). For example,
\eq{
  \tau_{\mc{S}_{\gamma}}(t) = \sum_{i=1}^{k} \tau_{\Omega_{i}}(t).
}
The quantity $\tau_{\mc{X}}(t) \in [0, 1]$ has the same physical units as the domain of the trajectory (e.g. of time, or perhaps length), hence we refer to the measurement $\tau_{\mc{S}_{\gamma}}$ as the \textit{loop duration} of the trajectory at time $t$. Setting the lower limit of integration on the integral in Eq.~\eqref{eq:loop_duration} to $t$ includes only inexact loops that begin at $\xi(t)$, while setting the upper limit of integration to $t$ includes only loops that end there. Considered on the whole set $\mc{X}=\mc{S}_{\gamma}$, a loop duration $\tau_{\mc{S}_{\gamma}}$ near zero implies that no other points on $\xi([0, 1])$ are nearby $\xi(t)$, while a loop duration of one implies that all of $\xi([0, 1])$ is in the vicinity of $\xi(t)$.

Integrating loop duration once again over a segment of a trajectory starting from a time $a$ and ending at a time $b$ gives us a measure of the total number of inexact loops that either begin or end along that segment, restricted to the set $\mc{X}$. 
\eq{
  \alpha_{\mc{X}}(a, b) &= \int_{a}^{b} \tau_{\mc{X}}(t) \, dt = \int_{a}^{b} \int_{0}^{1} \mbf{1}_{\mc{X}}(t, t') \, dt' \, dt \\
            &= \oiint_{\mc{A}} \, dt' \, dt, \, \text{with} \, \mc{A} = \mc{X} \cap [a, b] \times [0, 1].
%            &= \oiint_{\mc{} \cap [a, b] \times [0, 1]} \, dt' \, dt.
    \label{eq:loop_mass}
}
This surface integral leaves us with a simple mental picture for computing the quantity $\alpha_{\mc{X}}(a, b) \in [0, 1]$: it is the area of the set $\mc{X}$ between $t=a$ and $t=b$ in a trajectory's pairwise distance plot. Like loop duration, setting $\mc{X}=\mc{S}_{\gamma}$ gives a measurement of the area of all loop components, while $\mc{X}=\Omega$ measures only the area of a single loop component. Since the physical quantity time$^{2}$ has no name, we refer to $\alpha_{\mc{X}}(a, b)$ as the \textit{loop area} of the trajectory along the segment $\xi([a, b])$. The same argument of setting limits of integration to $t$ can be made for loop area as well, restricting the measurement to include only inexact loops that either begin or end on the segment.

Finally, we can define a trajectory segment's \textit{loop density} by averaging its loop area over the segment duration:
\eq{
\rho_{\mc{X}}(a, b) &= \frac{\alpha_{\mc{X}}(a, b)}{b - a}.
}
Loosely speaking, for $\mc{X}=\mc{S}_{\gamma}$, this final measurement tells us the average number of inexact loops attached to each point on the segment. Put more concretely, it measures the average loop duration of points on the segment. Loop area and loop density can be measured over the entire trajectory, in which case they are equivalent to one another in value, but have different units (since $\forall \mc{X} \in 2^{[0, 1]^2}, \, \rho_{\mc{X}}(0, 1) = \alpha_{\mc{X}}(0, 1)$).

%Table~\ref{table:concepts} summarizes the terminology and concepts related to inexact loops introduced throughout the previous sections.

\section{Sampling Inexact Loops}
\label{sec:sampling}

Armed with a definition for inexact loops and several tools for measuring them, we turn our attention to the more practical problem of building a representative sample of inexact loops from a provided trajectory. This problem finds relevance in SLAM and state estimation, where the goal is to reconstruct the trajectory taken by a system through its state space, given a set of noisy measurements constraining parts of the state. Sampling is important in this context because the number of inexact loops in a trajectory grows quadratically with time, i.e. in proportion to the loop area $\alpha_{\mc{S}_{\gamma}}(0, 1)$. A two hour long trajectory sampled at 10 Hz has close to three billion unique unordered pairs of sample points. Typical matching algorithms for generating constraints from a pair of sensor measurements, such as PnP~\cite{horaud1989analytic}, point cloud registration~\cite{segal2009generalized,yang2020teaser}, or specialized neural networks~\cite{chen2020overlapnet,zhang2017loop}, run on the order of tens to hundreds of milliseconds. Optimistically, densely matching all pairs would require close to one year on one CPU core.

\subsection{Sampling Complexity Classes}
\begin{figure}[t]
     \centering
     \begin{subfigure}[b]{0.48\textwidth}
         \centering
         \includegraphics[width=\textwidth]{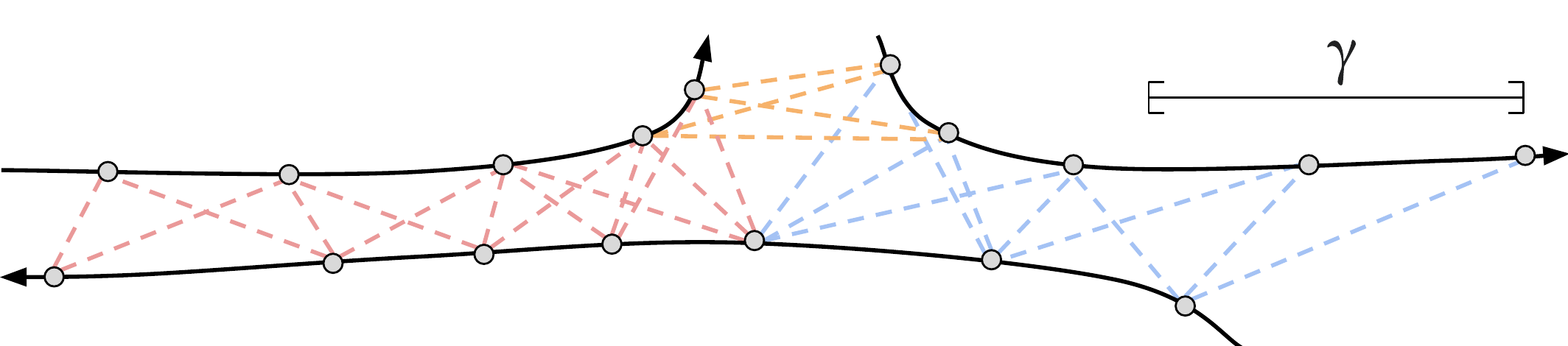}
         \caption{$\sigma \in \mc{O}(\alpha)$, \# samples / component / point $\propto \tau_{\Omega}(t)$.\label{fig:quadratic_sampling}}
     \end{subfigure}
     \\
     \begin{subfigure}[b]{0.48\textwidth}
         \centering
         \includegraphics[width=\textwidth]{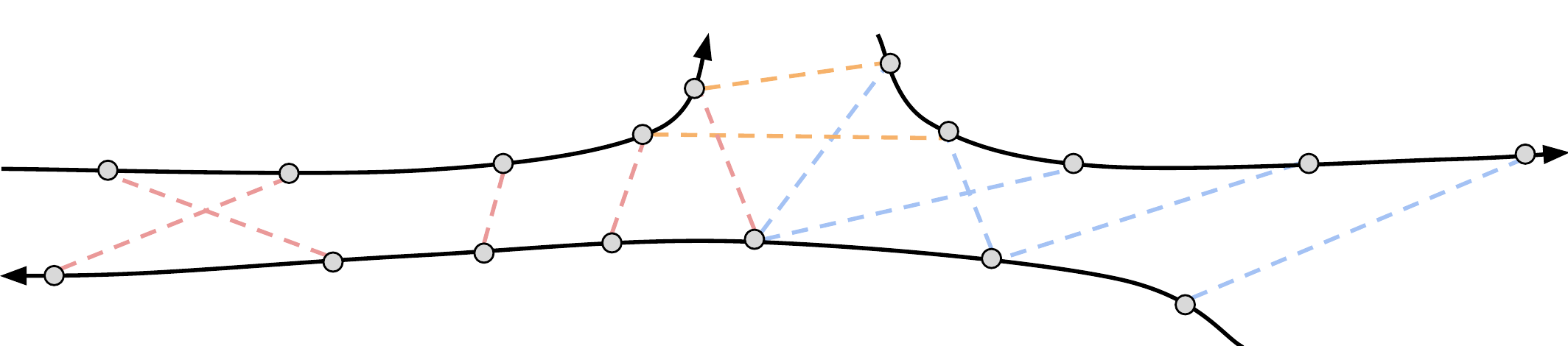}
         \caption{$\sigma \in \mc{O}(\rho)$, \# samples / component / point $\propto 1$.\label{fig:linear_sampling}}
     \end{subfigure}
     \\
     \begin{subfigure}[b]{0.48\textwidth}
         \centering
         \includegraphics[width=\textwidth]{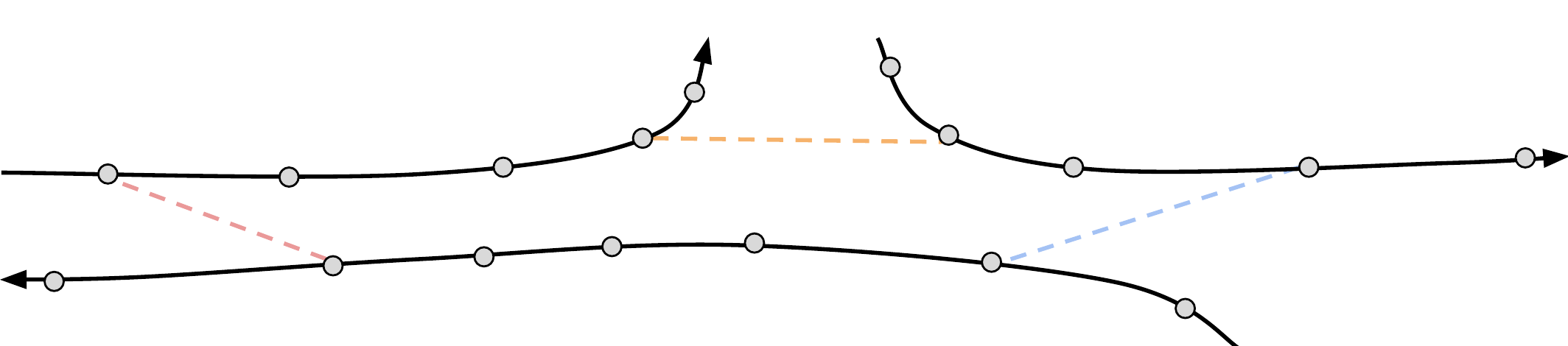}
         \caption{$\sigma \in \mc{O}(1)$, \# samples / component $\propto 1$.\label{fig:constant_sampling}}
     \end{subfigure}
     \caption{Possible complexity classes for a loop sampling algorithm $\sigma$. Samples from three non-trivial loop components of a trajectory are displayed (\textcolor{brick_red}{red}, \textcolor{burnt_orange}{orange}, \textcolor{royal_blue}{blue}).\label{fig:sampling_complexity}}
\end{figure}

For the moment we sidestep discussion of \textit{detection} of inexact loops, and consider sampling a representative subset as if we have the entire set $\mc{S}_{\gamma}$ available. It is possible to imagine three distinct complexity classes of inexact loop sampling algorithms. These correspond to whether the number of inexact loops sampled is proportional to (a) loop area, a measure of time$^2$, (b) loop density, a measure of time, or (c) the discrete number of loop components, a dimensionless measure (Fig.~\ref{fig:sampling_complexity}). Let $\Psi = \{l_1, l_2, \dots\} \subset S_{\gamma}$ be a finite sample of inexact loops beginning or ending on the segment $\xi([a, b])$. Let $\kappa \le k$ be the number of loop components intersecting $[a, b] \times [0, 1]$. These complexity classes yield the following numbers of samples: 
\begin{align}
\# \Psi &\propto \sum\nolimits_{i=1}^{\kappa} \alpha_{\Omega_{i}}(a, b) = \alpha_{\mc{S}_{\gamma}}(a, b) \tag{quadratic} \\
\# \Psi &\propto \sum\nolimits_{i=1}^{\kappa} \rho_{\Omega_{i}}(a, b) = \rho_{\mc{S}_{\gamma}}(a, b) \tag{linear} \\
\# \Psi &\propto \sum\nolimits_{i=1}^{\kappa} 1 = \kappa \tag{constant}.
\end{align}
If $\sigma : \mc{S}_{\gamma} \mapsto \Psi$ is a sampling algorithm, we say $\sigma \in \mc{O}(\alpha)$, $\sigma \in \mc{O}(\rho)$, or $\sigma \in \mc{O}(1)$ and refer to these as quadratic, linear, and constant complexity classes, respectively. It can be simpler to think of the classes $\mc{O}(\alpha)$ and $\mc{O}(\rho)$ in terms of the number of inexact loops that they produce per loop component $\Omega$, per sample time $t$. With this restriction, $\sigma \in \mc{O}(\alpha)$ generates samples in proportion to the loop duration $\tau_{\Omega}(t)$, while $\sigma \in \mc{O}(\rho)$ generates samples in proportion to a constant factor.

We remark that most existing loop closing strategies in SLAM literature belong to $\mc{O}(\alpha)$, including intelligent strategies that cluster detections into sets resembling loop components~\cite{olson2009recognizing,latif2013robust}, which still keep a number of constraints proportional to the loop area of each component.

% Mention that some keyframing techniques may be linear time!

% Fig: Sampling --------------------------------------------------------------

\subsection{Representative Samples}

What properties should a good inexact loop sampling algorithm exhibit? When is a sample $\Psi$ representative of the inexact loops on a trajectory? We propose the following desirable properties for a sampling algorithm $\sigma$:
\begin{enumerate}
\item [\textbf{P1}] Each unique loop component should be represented.
\item [\textbf{P2}] Similar inexact loops should be sampled infrequently.
\item [\textbf{P3}] Larger loop components should be sampled more often.
\end{enumerate}
\textbf{P1} is useful because we would prefer not to miss any locations where the trajectory returns to itself. This property is nearly a requirement in SLAM problems, because as previously mentioned, false negative loop closure detections can lead to arbitrarily large errors in a reconstructed trajectory. It is beneficial to err on the side of high recall than high precision~\cite{fawcett2006introduction}, since most modern SLAM back ends are robust against the presence of large fractions of outlier constraints~\cite{cadena2016past,tzoumas2019outlier,olson2013inference}. \textbf{P2} (i.e. sample dispersion) is desirable of any spatial sampling algorithm~\cite{ripley1979tests}. Again in the context of SLAM, building two loop constraints that convey the same information violates the fundamental assumption of independence of measurements~\cite{thrun2002probabilistic}. \textbf{P3} posits that large loop components generated when a trajectory remains near itself for a long period should be sampled more frequently than small loop components, where a trajectory nears itself only for a brief period.

Relating these desirable properties to the three sampling complexity classes, we see that $\textbf{P3}$ diminishes the usefulness of algorithms in $\mc{O}(1)$, where inexact loops are sampled independently of the size of the component they belong to. Algorithms in $\mc{O}(1)$ leave open the possibility of omitting important inexact loops in large loop components. These algorithms may still be useful in the case where one expects loop components to have similar areas. We also see that while all three complexity classes satisfy $\textbf{P1}$, common strategies such as random downsampling can violate this property. $\textbf{P1}$ asserts that regardless of algorithm, one must always cluster detected inexact loops into loop components and keep at least one sample per component, in order to avoid false negatives. This key algorithmic step is not present in many existing loop closing strategies. Finally, we see that $\textbf{P2}$ and $\textbf{P3}$ offer a trade off: if larger loop components are sampled more often, we obtain many inexact loops that convey similar information. Algorithms in $\mc{O}(\rho)$ correspond to a preference for $\textbf{P2}$, because although they sample more in larger loop components, this growth is sublinear in loop area. Algorithms in $\mc{O}(\alpha)$ sample in linear proportion to loop area, preferring $\textbf{P3}$.

\begin{figure*}[ht!]
     \centering
     \begin{subfigure}[b]{0.48\textwidth}
         \centering
         \includegraphics[width=0.3\textwidth]{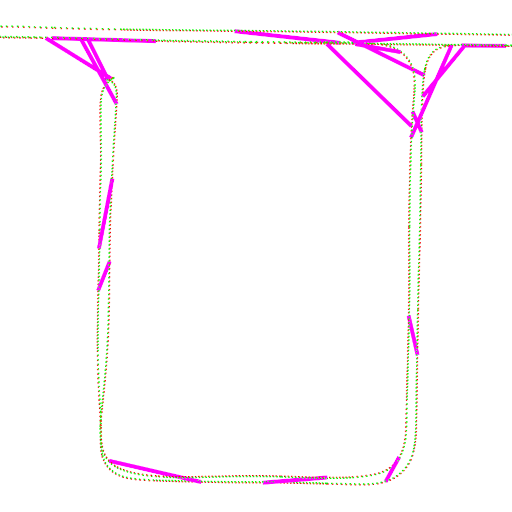}
         \hfill
         \includegraphics[width=0.3\textwidth]{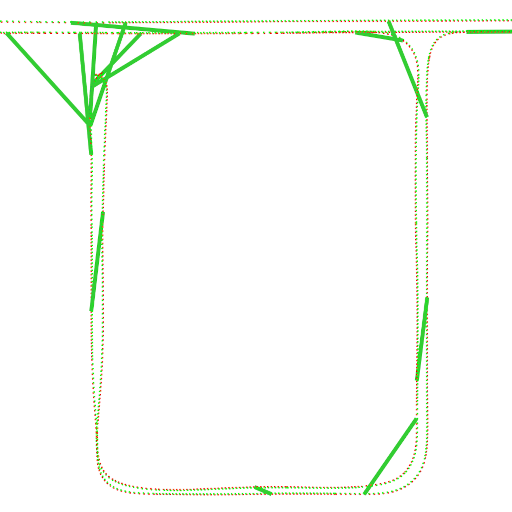}
         \hfill
         \includegraphics[width=0.3\textwidth]{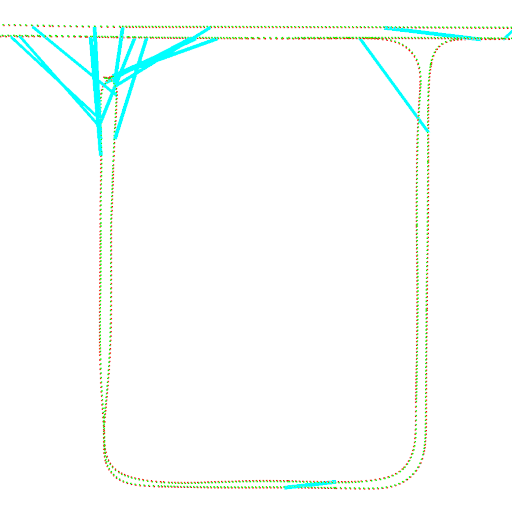}
         \caption{Samples from loop components with large loop areas.\label{fig:graph_large_loop_components}}
     \end{subfigure} 
     \hfill
     \begin{subfigure}[b]{0.48\textwidth}
         \centering
         \includegraphics[width=0.3\textwidth]{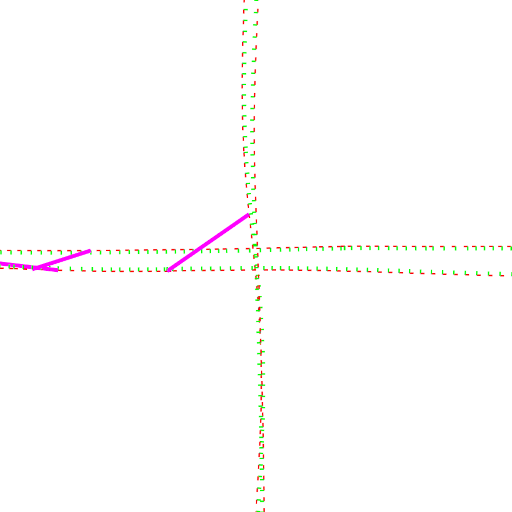}
         \hfill
         \includegraphics[width=0.3\textwidth]{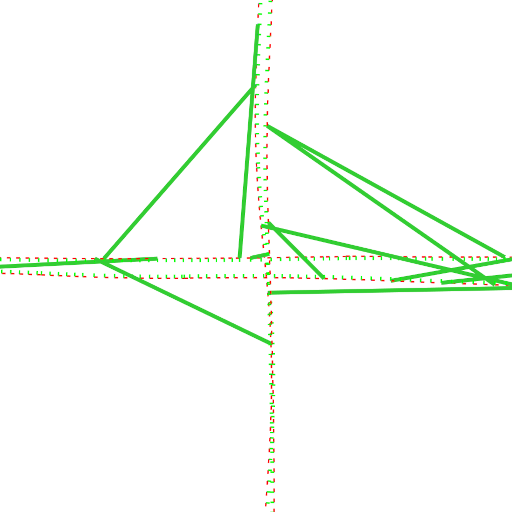}
         \hfill
         \includegraphics[width=0.3\textwidth]{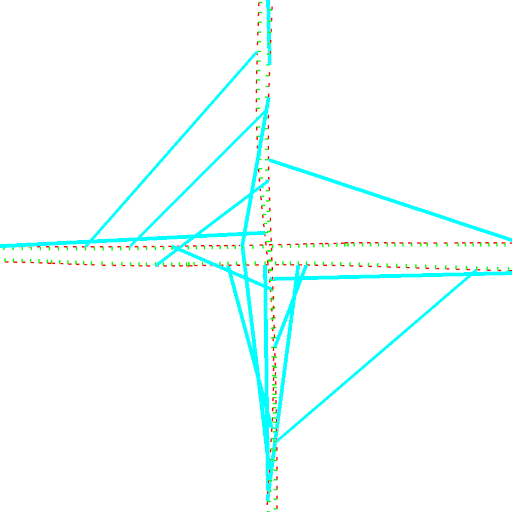}
         \caption{Samples from loop components with small loop areas.\label{fig:graph_small_loop_components}}
     \end{subfigure} 
     \\
     \begin{subfigure}[b]{0.69\textwidth}
         \centering
         \includegraphics[width=0.32\textwidth]{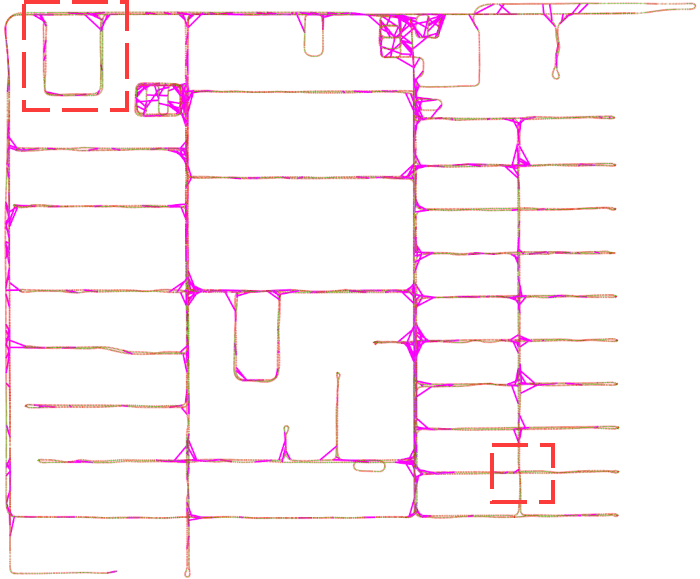}
         \hfill
         \includegraphics[width=0.32\textwidth]{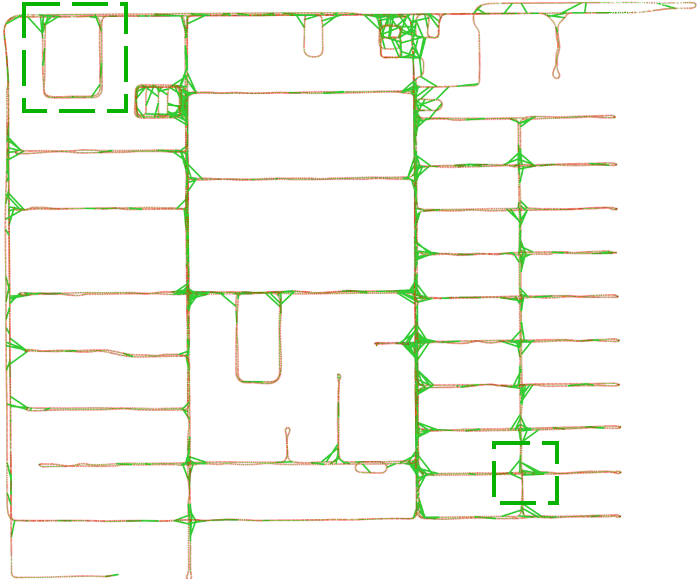}
         \hfill
         \includegraphics[width=0.32\textwidth]{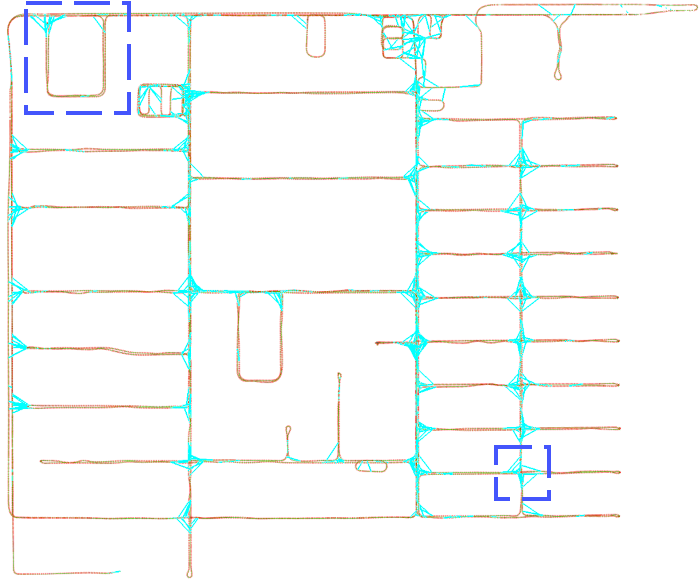}
         \caption{Samples from \textcolor{magenta}{$\sigma_{\alpha}$}, \textcolor{forest_green}{$\sigma_{\rho}$}, and \textcolor{aqua}{$\sigma_{1}$}, drawn between points on $\xi(t)$.\label{fig:graph_constant}}
     \end{subfigure}
     \hfill
     \begin{subfigure}[b]{0.27\textwidth}
         \centering
         \includegraphics[width=\textwidth]{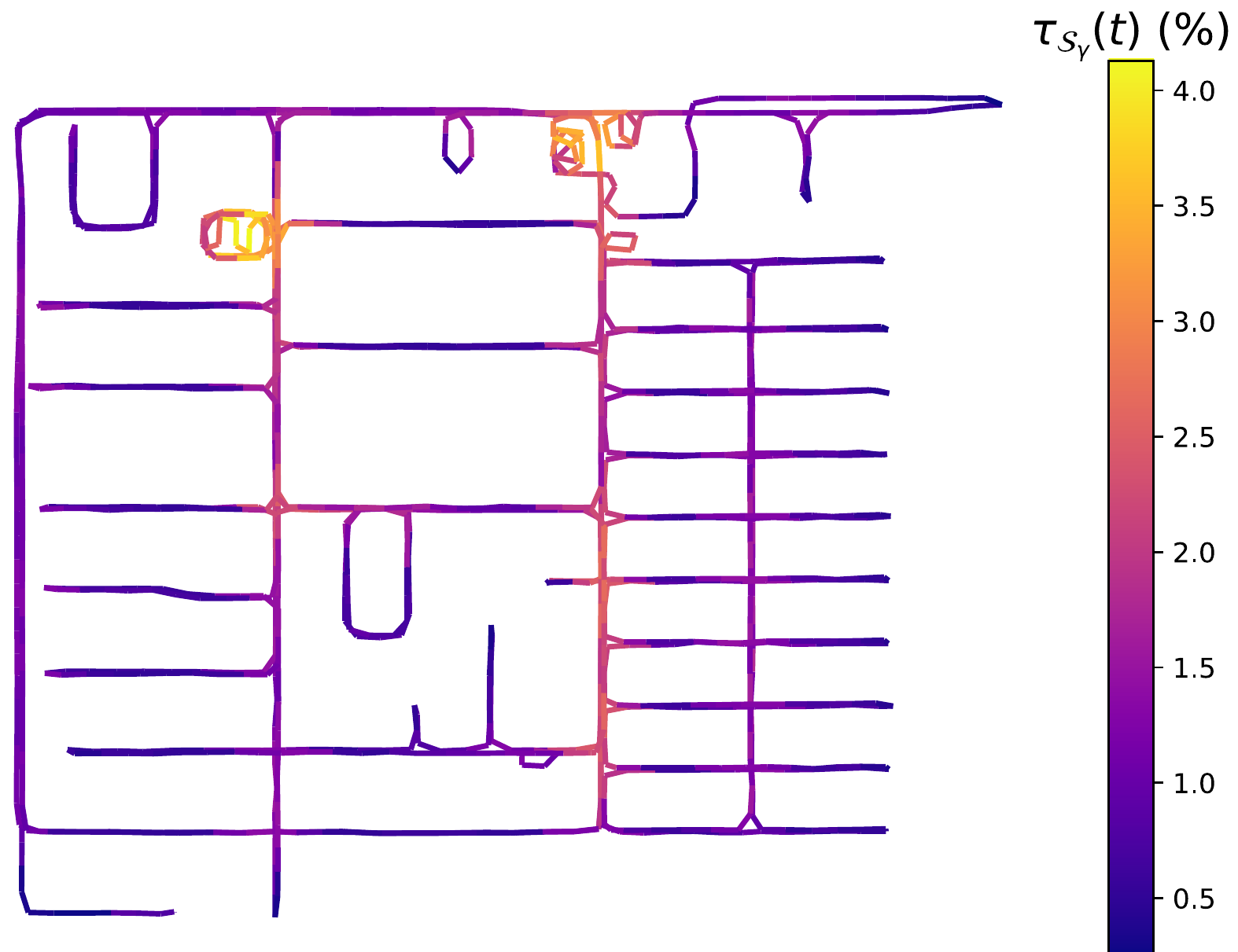}
         \caption{$\xi(t)$ colored by loop duration.\label{fig:graph_duration}}
     \end{subfigure}
     \caption{A $2.1$ hour trajectory through an urban area ($\mathtt{\sim}1$ km$^2$), with inexact loops sampled by algorithms from three complexity classes. Boxes highlight regions containing loop components with relatively high (Fig.~\ref{fig:graph_large_loop_components}) and low (Fig.~\ref{fig:graph_small_loop_components}) loop areas.\label{fig:graph}}
\end{figure*}

\subsection{Sampling From Detections}

To avoid false negatives, a sampling algorithm must necessarily identify loop components from its input. In the case that the input to $\sigma$ is the full set $\mc{S}_{\gamma}$, detecting loop components amounts to finding connected components on a regular 2D lattice, where vertices represent inexact loops of the form $(t, t')$, and edges are built in a $4$-neighborhood of each vertex. However, in many of the problem settings mentioned above, we are given a set of detections $\mc{D} = \{l_1, l_2, \cdots\} \subset \mc{S}_{\gamma}$ to sample from, rather than $\mc{S}_{\gamma}$ itself (i.e. $\Psi \subset \mc{D} \subset \mc{S}_{\gamma}$). In this case, we build a graph whose vertices are detected inexact loops $l = (t, t')$, and build edges between vertices whose $\mc{L}_{2}$ distance is less than a threshold duration $\epsilon$. The connected components on this graph form a suitable approximation of the trajectory's loop components. When $\mc{D}$ is a sparse subset of $\mc{S}_{\gamma}$, a larger $\epsilon$ will group more detections.

\subsection{Detecting Inexact Loops}

We briefly return to the topic of detecting inexact loops on a trajectory in absence of a priori knowledge of $\mc{S}_{\gamma}$, surveying existing solutions in order of hardness. Given a perfectly estimated trajectory (e.g. when planning simulated trajectories), detecting inexact loops simply involves caching and querying for neighboring points within $\gamma$. These operations are supported in most $O(1)$~\cite{lefebvre2006perfect,museth2013openvdb} or $O(n \log n)$~\cite{muja2009fast} spatial indexes, which sometimes support metrics on curved manifolds~\cite{sample2010tile,blanco2014nanoflann}. In this problem setting it is possible to build a discrete sample of the set $\mc{S}_{\gamma}$ with no approximation.

In offline state estimation tasks where sensor measurements that constrain part of the state in an absolute sense are available, detecting inexact loops involves first building an initial erroneous reconstruction of the trajectory, and then performing the same steps as above, perhaps with an inflated $\gamma$ to account for error. This case is applicable to settings such as warehouse robotics~\cite{beinschob2015graph} or mapping for autonomous vehicles~\cite{bresson2017simultaneous}, where measurements from GNSS, an IMU sensing gravity, sensors that match their measurements against an existing map~\cite{levinson2007map}, or detections of predeployed environment markers~\cite{wang2016apriltag} are available. Inexact loops detected in this manner are typically used afterwards to improve the estimate with additional constraints. In offline state estimation tasks where absolute measurements are not available, one must either increase $\gamma$ with time to account for expanding odometry error~\cite{rohou2018proving}, or find perceptually similar sensor measurements via techniques like visual hashing~\cite{yang2010bag} and place recognition~\cite{lowry2015visual}. The latter category amounts to using a pseudometric $\delta$ that measures distance in measurement space~\cite{di1999distance,glover2012openfabmap,wang2017deep}.

Inexact loop detection methods for online state estimation mirror their offline counterparts, except that their computation must be performed in real-time, fixing previous mistakes when new loops are detected so that the trajectory itself can be used for perception and planning. An additional tactic introduced in the online setting involves clustering detections~\cite{dong2015distributed} (sometimes into sets resembling loop components~\cite{olson2009recognizing,latif2013robust}), to reduce computational overhead.

\subsection{Sampling Experiments}

We conducted experiments using real trajectory data collected by a sensor-equipped robot on urban streets to compare benefits and drawbacks of sampling algorithms in the three separate complexity classes. Specifically, we wish to understand the behavior of algorithms in $\mc{O}(\alpha)$ in comparison to the other two classes. $\mc{O}(\alpha)$ contains most existing loop closing strategies, as well as the na\"{i}ve strategy of randomly sampling inexact loops from a set of detections with no concern for which loop component they fall in, which we hypothesize has the potential to both miss loop components, and over sample large loop components (violating \textbf{P1} and \textbf{P2}). 

Experiments were conducted in the offline state estimation setting with access to an approximately correct (up to 0.3 meters of absolute translation error) SLAM reconstruction of a $2.1$ hour long trajectory in $\mbb{SE}(3)$, with poses sampled at $10$ Hz. This led to approximately $7.6$e$4$ discrete timestamps from which inexact loops could begin or end. We began with implicit access to the entire $\mc{S}_{\gamma}$ by indexing and radius searching for poses in a KD-tree. Since this set is unwieldy (up to a possible $5.8$e$9$ elements, $\mathtt{\sim}86$ GB using 64 bit timestamps), we randomly sampled to $1\%$ to produce a set of detections $\mc{D}$, ensuring we retained enough that no loop components were omitted, and making the connected components algorithm tractable. We used the $\mc{L}_{2}$ metric on $\mathbb{R}^{3}$ pose translations, $\gamma = 40$ m, and $\epsilon = 30$ s. With these parameters the trajectory contained $490$ loop components of varying loop areas. For sampling algorithms we used $\sigma_{1} \in \mc{O}(1)$, which chose $c$ samples randomly from each loop component, $\sigma_{\rho} \in \mc{O}(\rho)$, which kept $r$ samples randomly per loop component per trajectory point, and $\sigma_{\alpha} \in \mc{O}(\alpha)$, which randomly sampled directly from $\mc{D}$, producing a number of samples proportional to $\alpha_{\mc{S}_{\gamma}}(0, 1)$. $\sigma_{1}$ and $\sigma_{\rho}$ intentionally kept at least one sample per loop component, while $\sigma_{\alpha}$ did not run connected components at all, and was therefore unable to.

We allowed each algorithm a budget of exactly $1$e$4$ inexact loop samples, forcing them to favor some loop components over others in relation to their complexity class. Fig.~\ref{fig:graph} shows the trajectory, as well as the inexact loops sampled by each algorithm. In regions containing loop components with large loop areas (Fig.~\ref{fig:graph_large_loop_components}), $\sigma_{1}$ under samples and omits relevant inexact loops. In regions containing loop components with small loop areas (Fig.~\ref{fig:graph_small_loop_components}), $\sigma_{\alpha}$ categorically fails to sample some components, leading to missed loop closures. $\sigma_{\rho}$ adequately samples inexact loops in both cases.

\begin{figure}[ht!]
     \centering
     \begin{subfigure}[b]{0.26\textwidth}
         \centering
         \includegraphics[width=\textwidth]{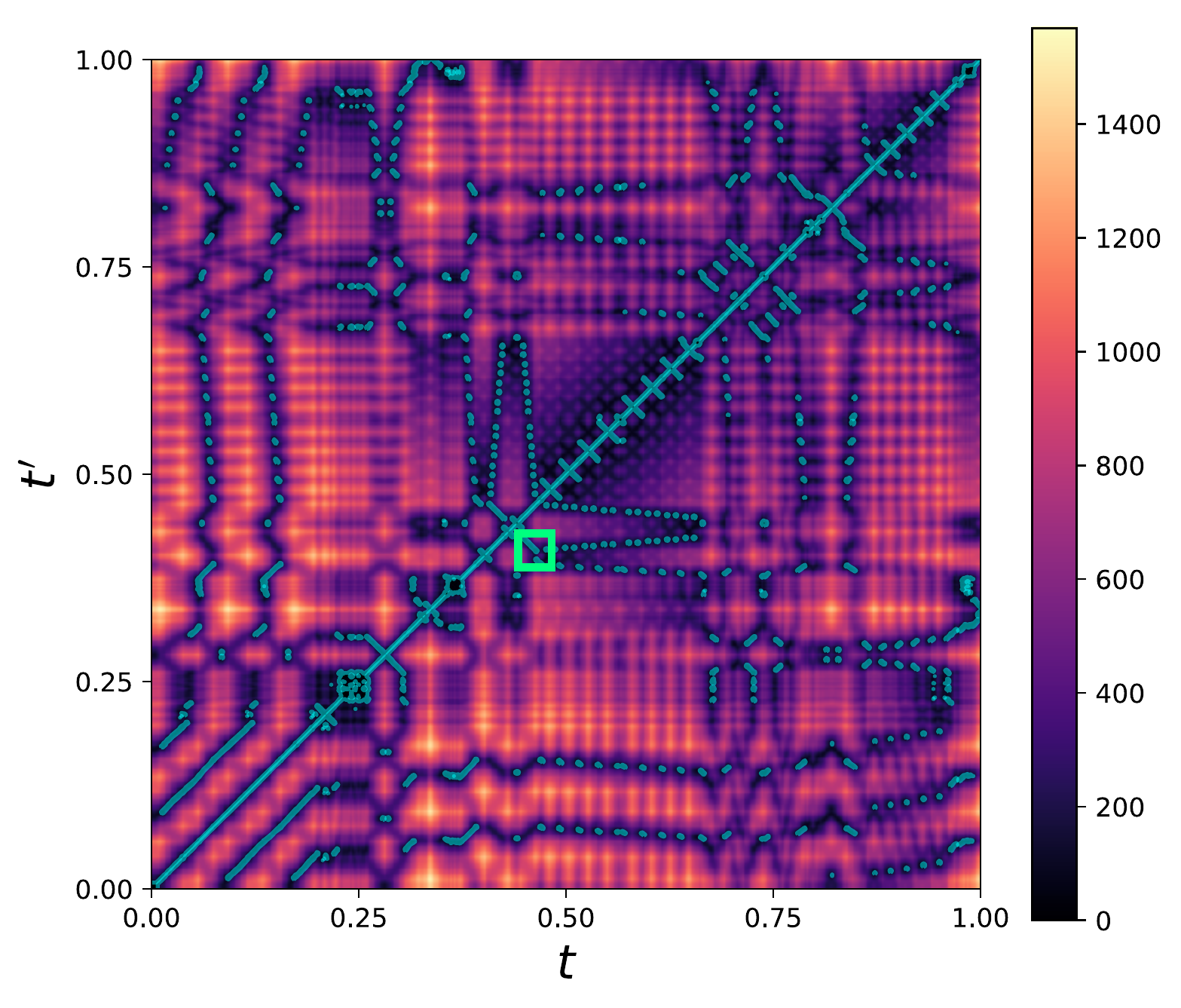}
         \caption{$\pi_{\xi}(t, t')$\label{fig:sample_scatter_dist}}
     \end{subfigure}
     \hfill
     \begin{subfigure}[b]{0.22\textwidth}
         \centering
         \includegraphics[width=\textwidth]{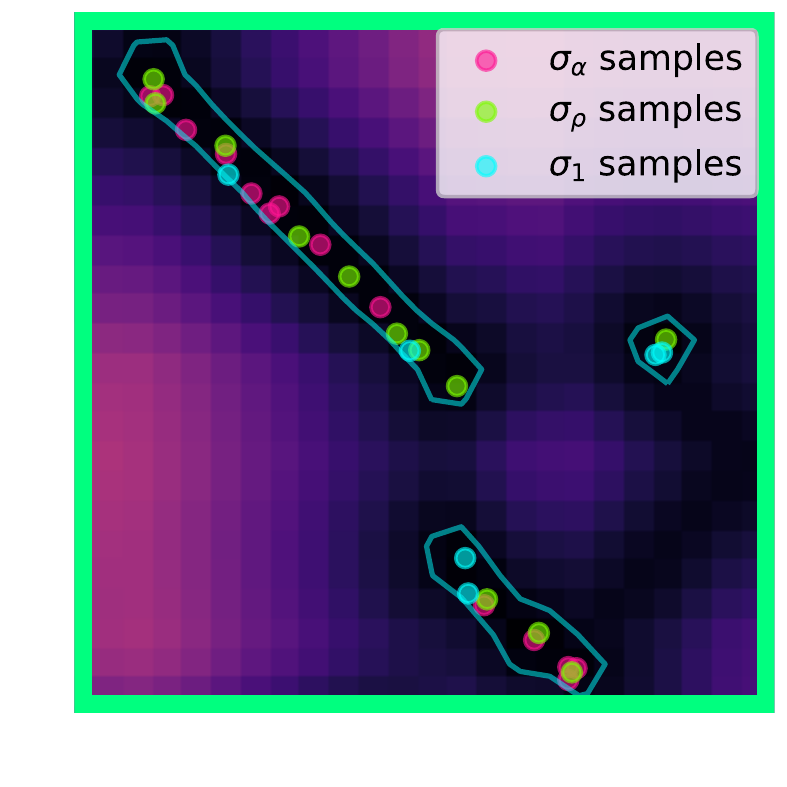}
         \caption{Sampled inexact loops.\label{fig:sample_scatter_zoom}}
     \end{subfigure} 
     \caption{Pairwise distance plot of experimental trajectory, and samples from \textcolor{magenta}{$\sigma_{\alpha}$}, \textcolor{forest_green}{$\sigma_{\rho}$}, and \textcolor{aqua}{$\sigma_{1}$} on three loop components with different areas ($\gamma=40$ m isocontours shown). Fig.~\ref{fig:sample_scatter_zoom} corresponds to the boxed region in Fig.~\ref{fig:sample_scatter_dist}.\label{fig:sample_scatter}}
\end{figure}

\begin{figure}[ht!]
     \centering
     \begin{subfigure}[b]{0.48\textwidth}
         \centering
         \includegraphics[width=\textwidth]{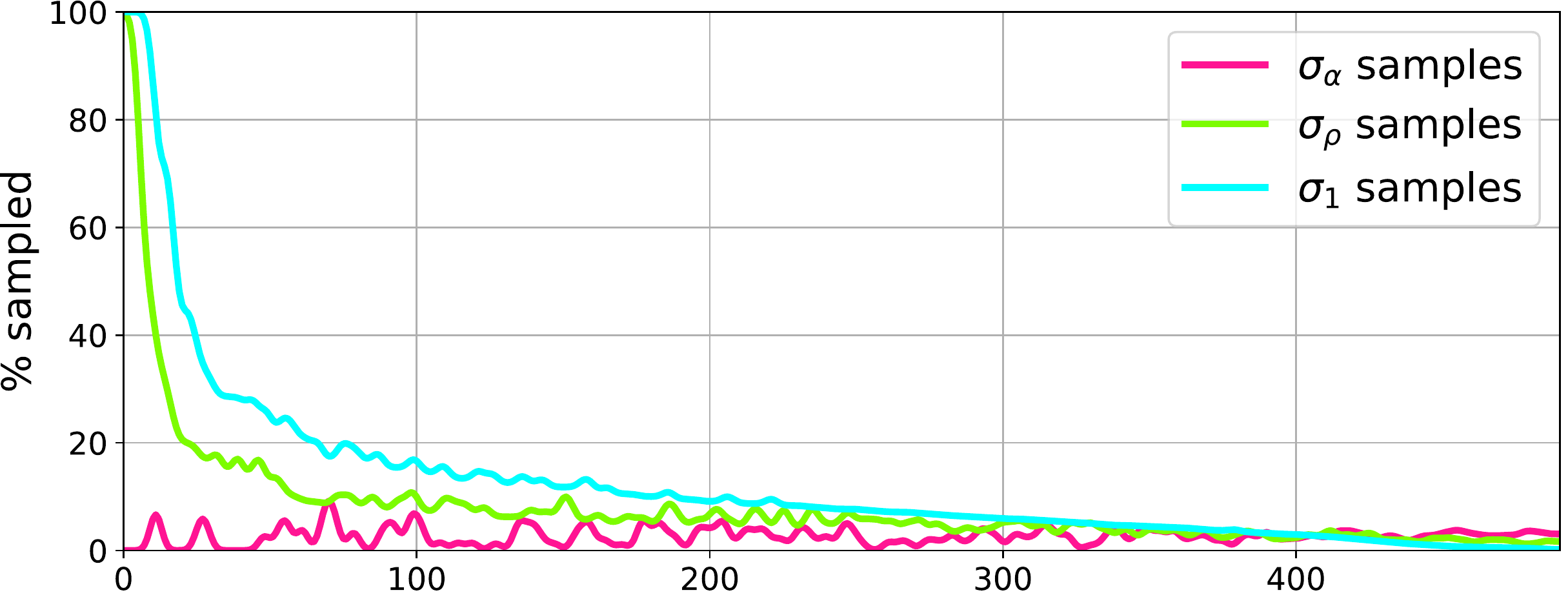}
     \end{subfigure}
     \\
     \begin{subfigure}[b]{0.48\textwidth}
         \centering
         \includegraphics[width=\textwidth]{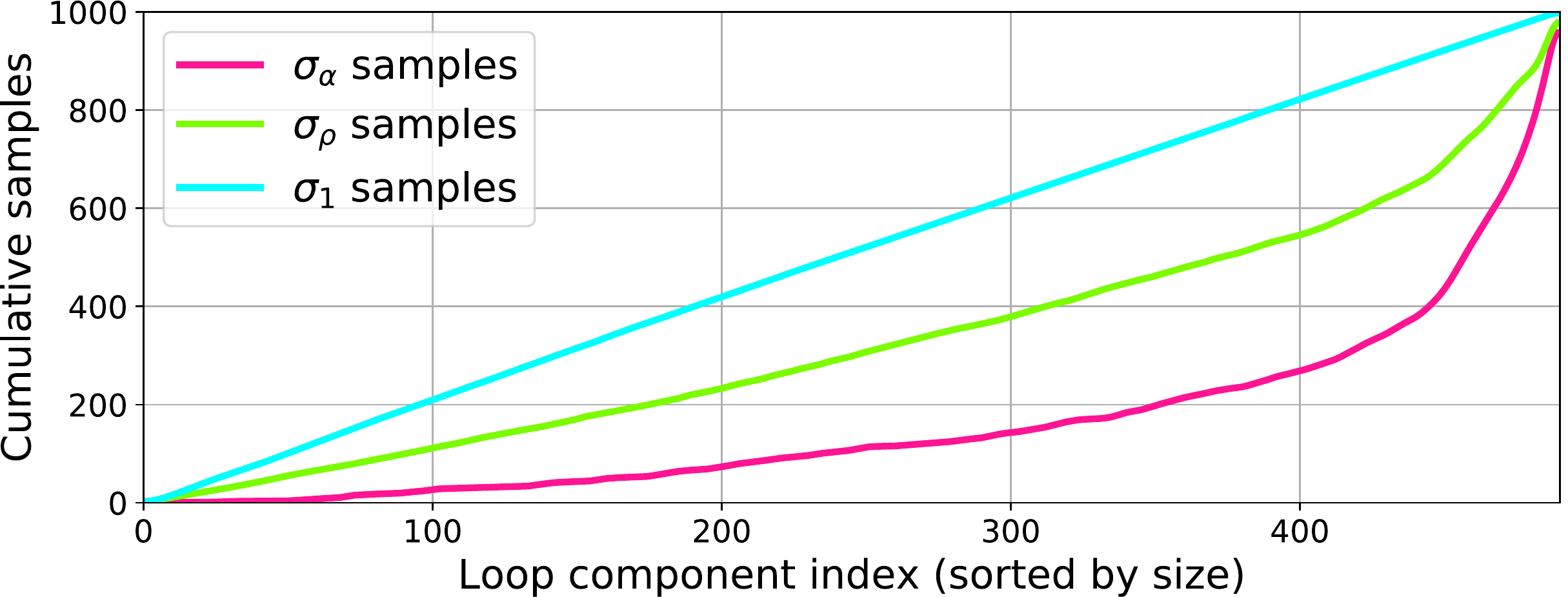}
     \end{subfigure} 
     \caption{Samples per loop component by algorithm.\label{fig:samples_versus_component_size}}
\end{figure}

The trajectory's pairwise distance plot is depicted in Fig.~\ref{fig:sample_scatter}, with loop components for $\gamma=40$ m outlined. A zoom in displays samples chosen by each algorithm in loop components of varying size, again revealing that $\sigma_{\alpha}$ omits small components and $\sigma_{1}$ under samples large ones (Fig.~\ref{fig:sample_scatter_zoom}). Fig.~\ref{fig:samples_versus_component_size} sorts loop components by their area, then shows the percentage (top) and cumulative number (bottom) of possible detections in $\mc{D}$ sampled from each component, showcasing the relationship between loop area and number of samples for each algorithm.

From these results, we readily suggest that other practitioners cluster inexact loops into loop components and retain at least one sample from each component, in order to avoid false negative loop detections. Furthermore, we suggest preferring sampling algorithms in the class $\mc{O}(\rho)$ that retain a fixed number of inexact loops per point, per component, as they sample more from loop components with larger areas, while avoiding over sampling in some high density areas. 

\section{Conclusion} 
\label{sec:conclusion}

In this work we formally introduced the concept of an inexact loop by relaxing the standard mathematical definition of a loop to allow for non-zero distance between its start and end points, which is of general use in robotics problems. We explored the topological consequences of this relaxed definition, including the existence and properties of path-connected clusters of inexact loops we called loop components. We then introduced several tools to measure inexact loops on a given trajectory. Finally, we characterized the complexity classes of inexact loop sampling algorithms, and compared several representative algorithms on long trajectories collected by a robotic vehicle on urban streets. 

We hope to see inexact loops and their related machinery find use broadly in robotics. Future work includes the application of inexact loops to the topic of motion planning, as well as a focused discussion on inexact loop sampling algorithms restricted to the difficult online SLAM setting.

\clearpage
\balance
\bibliographystyle{plainnat}

{\normalsize \bibliography{references}}
%{\small \bibliography{references}}

\end{document}